\documentclass{article}

\PassOptionsToPackage{numbers, compress}{natbib}

\usepackage[main, final]{neurips_2026}

\usepackage[utf8]{inputenc} 
\usepackage[T1]{fontenc}    
\usepackage{hyperref}       
\usepackage{url}            
\usepackage{booktabs}       
\usepackage{subcaption}
\usepackage{graphicx}
\usepackage{booktabs}
\usepackage{siunitx}
\usepackage{colortbl} 
\usepackage{amsmath}
\usepackage{cleveref}
\crefname{algocf}{algorithm}{algorithms}
\Crefname{algocf}{Algorithm}{Algorithms}
\usepackage{wrapfig}
\usepackage{enumitem}
\usepackage{array}
\usepackage{multirow}
\usepackage[ruled]{algorithm2e}
\usepackage{amsfonts}       
\usepackage{nicefrac}       
\usepackage{microtype}      
\usepackage{xcolor}         
\newcommand{\newtext}[1]{\textcolor{black}{#1}}
\newsavebox{\rfTableBox}
\newsavebox{\runtimeTableBox}
\newsavebox{\memoryTableBox}

\newcommand{\eg}{e.g.\ }
\newcommand{\ie}{i.e.\ }

\newtheorem{proposition}{Proposition}
\newtheorem{theorem}{Theorem}

\newcolumntype{C}{S[table-format=2.2]}
\newcolumntype{D}{S[table-format=2.2]}

\title{From Generic to Dedicated: A Novel Optimizer for Online Continual Learning}

\author{%
  Yongyi Wu\textsuperscript{*} \\
  University of Houston \\
  Houston, USA \\
  \texttt{yongyiwu.cn@gmail.com}
  \And
  Zheng Wang \\
  University of Houston \\
  Houston, USA \\
  \texttt{zwang214@cougarnet.uh.edu}
  \And
  Sen Lin\textsuperscript{\dag} \\
  University of Houston \\
  Houston, USA \\
  \texttt{slin50@central.uh.edu}
}

\begin{document}
\addtocontents{toc}{\protect\setcounter{tocdepth}{-1}}

\maketitle

\begingroup
\renewcommand{\thefootnote}{}
\makeatletter
\long\def\@makefntext#1{\noindent#1}
\makeatother
\footnotetext{\hspace*{1em}\textsuperscript{*}Work done during an internship at University of Houston\\
\hspace*{1em}\textsuperscript{\dag}Corresponding author: \texttt{slin50@central.uh.edu}}
\endgroup

\begin{abstract}
    Online Continual Learning (OCL) requires models to learn from non-stationary data streams under a strict single-pass constraint, making them highly susceptible to catastrophic forgetting. While existing studies have explored various strategies from data, architecture, and optimization perspectives, the optimizer often directly uses generic approaches such as SGD and AdamW. In this work, we reveal a new link between optimization dynamics and OCL needs by recasting Polyak-Ruppert averaging as the engine of ``plasticity'' and Primal averaging as the anchor of ``stability''. This inspires our novel optimizer, SPIN (\textbf{S}tability-\textbf{P}lasticity \textbf{IN}terpolation), which explicitly decouples plasticity and stability by interpolating between a fast-moving ``plasticity'' sequence and an adaptive ``stability'' sequence. We theoretically demonstrate that SPIN implicitly employs an inverse Hessian approximation, providing crucial Tikhonov regularization to damp noisy, single-sample Hessian estimates and mitigate forgetting. More importantly, SPIN can be seamlessly integrated into existing OCL methods by taking resulted gradients as input and replacing standard optimizers. Extensive experiments show that using SPIN instead of standard optimizers yields consistent and significant performance gains across multiple OCL approaches, including the challenging rehearsal-free setting and under strict memory constraints.
\end{abstract}

\section{Introduction}
\label{sec:intro}

Continual Learning (CL) aims to enable AI systems to continuously acquire new knowledge while retaining previously learned skills, a capability essential for many real-world applications  where data arrives sequentially~\cite{CLsurveyWang,CLsurveyZhou,vandeVen2019ThreeSF,DeSurvey}. However, the common assumption in offline CL,  that models are able to revisit task data multiple times during training, rarely holds. This has motivated the rise of Online Continual Learning (OCL), a more realistic paradigm where each sample is observed only once~\cite{He2020IncrementalLI,soutifcormerais2023improving,MAI202228}. Nevertheless, OCL inevitably exacerbates key challenges like catastrophic forgetting~\cite{french1999catastrophic,Kirkpatrick2016OvercomingCF} and the plasticity-stability dilemma~\cite{mermillod2013stability} under strict single-pass constraints.

To address this, existing OCL research has largely followed the classical offline CL paradigm, focusing on data, loss functions, and architectural designs, such as replay, regularization, and distillation~\cite{Rolnick2018ExperienceRF,Caccia2021NewIO,Huszr2017NoteOT,SI,yan2024orchestrate}. In most of these approaches, the lower-level optimizer is treated as a fixed, secondary component, directly using generic strategies such as vanilla SGD~\cite{SGD} or AdamW~\cite{Adam,AdamW}. In both SGD and AdamW, the gradient is evaluated at the current parameter given the newly arrived sample. Consequently, the optimization direction is dominated by the instantaneous data distribution, even when momentum is used. This behavior can lead to accelerated forgetting under non-stationary data streams. From the perspective of online optimization theory, however, the optimizer itself plays a decisive role in determining convergence, stability, and generalization~\cite{OnlineSurvey}.
Many works have explored the role of optimization in OCL, proposing methods that modify gradient updates via projection, reweighting, proximal constraints, or curvature-aware preconditioning~\cite{gem,agem,wumitigating,LPR}. These approaches operate primarily at the level of gradient manipulation, adapting the update direction while still relying on standard optimizers. Therefore, a critical question arises: ``\emph{Instead of relying on generic optimizers, can we design a novel optimizer specifically tailored to OCL that effectively mitigates the forgetting problem and further pushes the limits of existing OCL approaches?}''

\begin{wrapfigure}{r}{0.51\linewidth}
\vspace{-0.5cm}
  \centering
  \includegraphics[width=\linewidth]{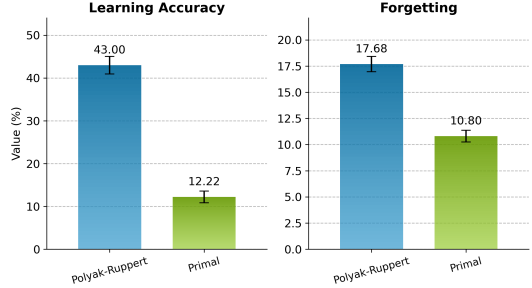}
  \caption{The comparison of plasticity (Learning Accuracy) and stability (Average Forgetting) of Polyak-Ruppert averaging and Primal averaging on split CIFAR-100. Learning Accuracy for the $j$-th task is the accuracy evaluated on the test set of task $j$ immediately after the network has been trained sequentially from task $1$ up to task $j$.}
  \label{fig:1}
\end{wrapfigure}
To answer this question, we revisit the fundamentals of stochastic optimization and establish a direct connection between its averaging mechanisms and the stability–plasticity dilemma in OCL. 
Specifically, we reinterpret \textit{Polyak–Ruppert averaging}~\cite{polyak,Ruppert} as the engine of \textbf{plasticity}, since it evaluates gradients at the current, fast-evolving iterate and thus quickly adapts to new data. In contrast, \textit{Primal averaging}~\cite{primal,primal2,gpa} serves as the anchor of \textbf{stability}, as it computes gradients at the averaged, slowly updating point, thereby retaining previously acquired knowledge. We empirically examine these two averaging schemes, as shown in \Cref{fig:1}. Polyak–Ruppert averaging achieves higher learning accuracy but suffers from greater forgetting, reflecting its strong plasticity yet limited stability. Primal averaging, on the other hand, exhibits the opposite behavior, showing lower plasticity while maintaining superior stability. 

Driven by these observations, we propose a novel optimizer, SPIN (\textbf{S}tability-\textbf{P}lasticity \textbf{IN}terpolation), which explicitly manages this trade-off by operating on two coupled iterate sequences. SPIN maintains a plasticity model sequence for rapid adaptation and a stability model sequence for consolidated knowledge. Its core mechanism links these sequences in two ways. First, gradients are evaluated at an interpolated point between the two models, enabling a unified view of multiple averaging strategies. Second, the stability sequence is updated by smoothly integrating information from the plastic sequence, preventing corruption under the high-variance, non-stationary OCL stream. We find that this formulation not only blends the benefits of different averaging approaches, but also connects conceptually to several regularization-based and meta-continual learning methods. Our main contributions can be summarized as follows:

(1) We propose a novel optimizer dedicated for OCL, namely SPIN, which explicitly decouples plasticity and stability at the model level. Crucially, SPIN can be directly integrated into existing OCL methods by simply replacing their lower-level optimizers. In other words, \emph{SPIN operates similarly to SGD by taking only the gradients as input, regardless of how those gradients are modified by the upstream OCL methods.}
    
(2) We establish a unified optimization framework for OCL, providing new theoretical insights into the plasticity-stability dilemma. Within this framework, we theoretically justify the design of SPIN and its effectiveness in mitigating catastrophic forgetting.
    
(3) Extensive experiments validate the effectiveness and generality of SPIN, showing consistent, significant performance gains across diverse OCL approaches, including the challenging rehearsal-free setting and under strict memory constraints.

\section{Related Work}
\label{sec:formatting}

\textbf{Continual Learning.} CL aims to enable models to learn from a sequence of tasks without forgetting previous knowledge~\cite{Ostapenko2019LearningTR,CLsurveyZhou,linicmltheory}. Conventionally, CL strategies primarily fall into three categories: 
Memory-based methods revisit past knowledge, including replay-based methods that retrain on stored old data~\cite{Rolnick2018ExperienceRF,derpp,WangTMM,dengiclrtheory} and orthogonal projection-based methods that project gradients to prevent interference~\cite{wumitigating,lin2022trgp,agem}; Regularization-based methods, which add penalties to the loss function to prevent significant changes to weights deemed important for old tasks~\cite{afec,Kirkpatrick2016OvercomingCF,zhoudbn}; and Parameter Isolation methods, which use dedicated network parameters for different tasks, either by expanding the architecture or by isolating pathways within a fixed network~\cite{wu2025sdlora,hidelora,inflora,Yan2021DERDE,liniclrtheory}. \newtext{CLS-ER combines short- and long-term semantic memories with experience replay to support adaptation and knowledge retention~\cite{arani2022learning}.} \newtext{Parameter averaging offers another approach: CoMA combines model parameters from previous and current tasks to balance stability and plasticity, while CoFiMA further weights parameters using Fisher information~\cite{marouf2024weighted}.} While optimization-oriented methods exist in offline CL~(e.g., \cite{Wang_2021_CVPR,lin2022trgp,GPM}), their fundamental reliance on accessing past data  makes them incompatible with the single-pass constraints of OCL.

 \textbf{Online Continual Learning.} OCL requiress models to learn from a strict single-pass data stream~\cite{OnlineSurvey}. Replay-based strategies dominate this space~\cite{soutif2023comprehensive}, ranging from simple buffer combinations (ER~\cite{Rolnick2018ExperienceRF}, DER~\cite{derpp}) to advanced objective modifications addressing representation drift (ER-ACE, OCM, GSA, OnPro~\cite{Caccia2021NewIO,ocm,onpro,GSA}). Recent methods further boost adaptability via multi-level supervision (MOSE~\cite{yan2024orchestrate}) or advanced state-space and collaborative modeling (CCL-DC~\cite{ccldc}, S6MOD~\cite{s6mod}). \newtext{CBA addresses task-recency bias under distribution shifts through a continual bias adaptor~\cite{wangcba}.} A-GEM constrains the current gradient to avoid increasing the loss on past data by projecting it onto the feasible region defined by reference gradients from memory~\cite{agem}. LPR applies a layer-wise proximal preconditioning that stabilizes optimization by limiting abrupt changes to past representations while learning from both new and replay data~\cite{LPR}. \emph{In contrast, our work focuses on the lower-level optimizer design that takes upper-level gradients as input, and provides a complementary and orthogonal perspective that can be applied on top of diverse OCL methods.}

\newtext{\textbf{Averaging-Based Optimization.} Classical optimization methods use coupled iterate sequences and averaging to accelerate convergence or reduce variance, as exemplified by Nesterov's accelerated methods and Polyak--Ruppert averaging~\cite{nesterov1983method,nesterov2018lectures,polyak,Ruppert}. Further analyses establish connections between iterate averaging, momentum, and convergence guarantees~\cite{Shamir2012StochasticGD,Defazio2020ThePO}. In deep learning, stochastic weight averaging (SWA) improves generalization by averaging weights along the training trajectory~\cite{izmailov2018averaging}, while Lookahead periodically interpolates and synchronizes fast and slow weights~\cite{lookahead}. Schedule-Free optimization combines interpolated gradient evaluation with cumulative iterate averaging to avoid a prespecified training horizon~\cite{defazio2024road}, with nonconvex convergence guarantees established for its SGD variant~\cite{ahn2025general}. Generalized Primal Averaging (GPA) uses independent coefficients for gradient evaluation and iterate averaging to smooth DiLoCo updates and improve LLM training efficiency~\cite{gpa}. SPIN shares this coupled interpolation-and-averaging structure. Our focus is on how the gradient evaluation point and averaging dynamics jointly influence the stability–plasticity trade-off in non-stationary OCL, where each update must accommodate incoming data while preserving knowledge of previously encountered tasks.}

\section{Preliminaries}

\textbf{Online Continual Learning Setting.}
Consider the standard OCL setting, where a model $f_\theta$ with parameters $\theta$ learns from a non-stationary data stream of $T$ tasks~\cite{He2020IncrementalLI,soutifcormerais2023improving,MAI202228}. The dataset for the $t$-th task is denoted as $\mathcal{D}_t = \{(v_i, y_i)\}$ with input-label pairs, where $\mathcal{C}_t$ is the set of classes for the $t$-th task. Let $\mathcal{D} = \bigcup_{t=1}^T \mathcal{D}_t$. Following the common  protocol that $\mathcal{C}_t \cap \mathcal{C}_{t'} = \emptyset$ for $t \neq t'$, the model must infer the correct class from the entire set of seen classes $\mathcal{Y} = \bigcup_{i=1}^T \mathcal{C}_i$ at test time~\cite{soutif2023comprehensive}. The model processes this data stream sequentially online, where data from each task  is provided as a stream of mini-batches. We denote a single mini-batch from the current stream as $\mathcal{B}_{\text{stream}} = \{(v_i, y_i)\}_{i=1}^B$, with batch size $B$. Crucially, OCL imposes a strict \textbf{single-pass constraint}: the model observes each training sample $(v, y)$ \textit{exactly once} and cannot re-access past task data (unless stored in a buffer). This constraint, combined with data distribution shift, leads to catastrophic forgetting~\cite{french1999catastrophic}, where the model's performance on previous tasks degrades significantly as it learns new ones. 
To mitigate forgetting, many OCL methods employ  experience replay~\cite{Rolnick2018ExperienceRF}. A small memory buffer $\mathcal{M}$ with a fixed capacity $M$  is maintained to store a subset of past samples. At each training step, the model trains on the union of the current stream batch and a replay batch drawn from $\mathcal{M}$. The goal is to learn  parameters $\theta_T$ at the end of the training stream that minimizes the empirical risk across all encountered tasks:
$\min_{\theta_T} \sum_{t=1}^T \mathbb{E}_{(v,y) \sim \mathcal{D}_t} \left[ \mathcal{L}(f(v; \theta_T), y) \right]$,
where $\mathcal{L}$ is the loss function.

\textbf{Averaging-Based Methods in Optimization.}
Nesterov's accelerated method is an optimization algorithm designed to achieve faster convergence rates for $L$-smooth convex functions~\cite{nesterov1983method,nesterov2018lectures}. One common form, known as  AC-SA~\cite{lan2012optimal}, utilizes an interpolation between two sequences of iterates:
\begin{align}
    x_t &= (1-c_{t+1})\theta_t + c_{t+1}\phi_t, \label{eq:nam_y} \\
    \phi_{t+1} &= \phi_t - \frac{t+1}{2L}\nabla \mathcal{L}(x_t), \label{eq:nam_z} \\
    \theta_{t+1} &= (1-c_{t+1})\theta_t + c_{t+1}\phi_{t+1}. \label{eq:nam_x}
\end{align}
Here, $\phi_t$ is the base sequence, $\theta_t$ is the averaged sequence, and $x_t$ is the point at which the gradient is evaluated. The coefficient $c_{t+1}$ is typically set as $c_{t+1} = 2/(t+2)$ for non-strongly convex functions.

\textbf{Gradient-Space Interpolation in OCL.}
Many replay-based OCL methods can be written as a gradient-space interpolation between the current task and past tasks~\cite{wumitigating}. Let $g_t^{\text{cur}}$ denote the gradient on the current sample at step $t$, and $\{g_t^{(i)}\}_{i < t}$ the gradients estimated from the replay buffer for previous tasks. A generic form is
\begin{equation}
\textstyle
    u_t \;=\; g_t^{\text{cur}} \;+\; \sum_{i < t} \lambda_i \, g_t^{(i)},
    \label{eq:grad_interp}
\end{equation}
where $u_t$ is the update direction fed to lower-level optimizers and $\lambda_i$ are task-dependent weights.

Different choices of $\lambda_i$ recover many existing OCL algorithms: experience replay methods such as ER and DER use uniform averaging over buffered gradients~\cite{Rolnick2018ExperienceRF,derpp,wumitigating}; gradient-alignment methods such as GEM, A-GEM, and MER implicitly realize Eq.~\ref{eq:grad_interp} via projection-based or bi-level updates~\cite{gem,agem,MER}; Pareto-style methods (\eg, POCL) optimize $\lambda$ to balance previous tasks within a multi-objective formulation~\cite{wumitigating}. In all these cases, the gradients are evaluated at the current parameter $\theta_t$, and only their linear combination in gradient space is modified.

\textbf{Which Averaging Mechanism Suits OCL.}
Under the viewpoint of Polyak–Ruppert averaging and Primal averaging, the two possible ways to form an averaged direction are: (i) averaging the gradients at $\phi_t$ and $\theta_t$, or (ii) averaging the iterates and then taking a single gradient at the interpolated point.
Gradient-space averaging's formulation can be generalized as a weighted average of gradients:
\begin{equation}
    g_t^{\text{grad}}
    = (1-\beta)\nabla \mathcal{L}(\phi_t)
    + \beta\nabla \mathcal{L}(\theta_t).
    \label{eq:grad_avg}
\end{equation}
Compared with directly averaging gradients from different iterates, parameter-space averaging evaluates the gradient at a single interpolated point between two coupled sequences:
\begin{equation}
    g_t^{\text{param}}
    = \nabla \mathcal{L}\!\left( (1-\beta)\phi_t + \beta\theta_t \right).
\end{equation}
This formulation induces a coherent optimization direction along the shared parameter manifold while implicitly incorporating local curvature information through the interpolated evaluation point. In contrast, gradient-space averaging mixes gradients computed at different locations in parameter space, which may lead to inconsistent update directions under non-stationary data streams. Motivated by these observations, we adopt parameter-space interpolation as the core design principle of SPIN. In ~\Cref{sec4.2}, we further show that this formulation admits a unified optimization interpretation and naturally connects to curvature-aware continual learning methods. Empirical comparisons between gradient-space and parameter-space averaging are presented in ~\Cref{ablation}.

\section{Method}
In this section, we first present the general form of our SPIN optimizer in \Cref{sec4.1}, and then provide a theoretical perspective that connects our optimizer with existing methods in \Cref{sec4.2}. Finally, we establish the theoretical guarantees and empirical validation of our optimizer in \Cref{sec4.3}.

\subsection{The SPIN Optimizer}\label{sec4.1}

We propose our SPIN optimizer as follows:

\begin{align}
    x_t &= (1-\beta)\phi_t + \beta \theta_t \label{eq:fsm_y} \\ 
    \phi_{t+1} &= \textit{Opt}(\phi_t, \nabla \mathcal{L}(x_t), \gamma) \label{eq:fsm_z}\\ 
    \theta_{t+1} &= \theta_t + \alpha(\phi_{t+1} - \theta_t) \label{eq:fsm_x} 
\end{align}

where $\textit{Opt}(\cdot)$ denotes the update function of a standard base optimizer (\eg, SGD or AdamW) with learning rate $\gamma$. Here, $\phi_t$ is the fast-moving plasticity sequence, $x_t$ is the interpolated point where the stochastic gradient $\nabla \mathcal{L}(x_t)$ is evaluated, and $\theta_t$ is a slow-moving stability sequence that evolves jointly. The coefficients $\alpha, \beta \in [0,1]$ control the synchronization strength and the interpolation ratio between the two sequences, respectively. Note that $\mathcal{L}$ is non-stationary due to the changing data distributions~\cite{soutif2023comprehensive,MAI202228} and depends on the upstream algorithm design. Detailed pseudo-code of the SPIN algorithm is provided in Appendix~\ref{pcode}.

SPIN adopts a synchronous dual-sequence coupling mechanism: the plasticity sequence $\phi_t$ rapidly adapts to incoming data, while the stability sequence $\theta_t$ incrementally consolidates these updates into a robust representation. This design ensures that both sequences evolve jointly at every iteration. The plasticity sequence $\phi_t$ captures fast adaptation to new data, whereas the stability sequence $\theta_t$ gradually consolidates these changes into a consistent and robust representation. In the OCL setting, where the data distribution drifts over time, such tightly coupled dynamics allow $\theta_t$ to promptly absorb relevant new information while preventing abrupt forgetting of past knowledge. Moreover, SPIN evaluates the gradient at the interpolated point $x_t$. This intermediate evaluation allows the optimizer to blend the short-term plastic response of $\phi_t$ with the long-term stability encoded in $\theta_t$, effectively guiding the update toward a compromise direction between adaptation and retention.

\subsection{A Unified View}\label{sec4.2}
To understand the internal dynamics of the SPIN optimizer, we first analyze its update rules from first principles. The optimizer maintains two sequences, $\phi_t$ and $\theta_t$, updated as per Eq.~\ref{eq:fsm_x}. Gradients $g_t = \nabla \mathcal{L}(x_t)$ are computed at $x_t = \theta_t + (1-\beta)d_t$, where $d_t = \phi_t - \theta_t$ is the gap vector. First, assuming the loss function $\mathcal{L}_t(\theta)$ is sufficiently smooth, we employ the first-order Taylor expansion of the gradient $g_t = \nabla \mathcal{L}(x_t)$ around the iterate $\theta_t$, which gives:
\begin{equation} \label{eq:taylor_expansion_gt}
\begin{aligned}
g_t \approx \nabla \mathcal{L}_t(\theta_t) + \nabla^2 \mathcal{L}_t(\theta_t) (x_t - \theta_t) 
    = \nabla \mathcal{L}_t(\theta_t) + (1-\beta) H_t d_t
\end{aligned}
\end{equation}
where $H_t := \nabla^2 \mathcal{L}_t(\theta_t)$ is the Hessian at $\theta_t$. 
Second, invoking a steady-state assumption for the gap vector $d_t$ (\ie, $d_{t+1} \approx d_t$), allows us to express $d_t$ as a function of the gradient $\nabla \mathcal{L}_t(\theta_t)$ and Hessian $H_t$~\cite{qss2}. This analysis leads to the following result:
\begin{proposition}Suppose the gradient $g_t = \nabla \mathcal{L}(x_t)$ can be approximated by its first-order Taylor expansion around $\theta_t$ as in Eq.~\ref{eq:taylor_expansion_gt}, and assume the gap vector $d_t = \phi_t - \theta_t$ reaches a quasi-steady state ($d_{t+1} \approx d_t$). Then the effective update for the iterate $\theta_t$ simplifies to:
\begin{equation} \label{eq:sf_effective_update_final_revised}
\theta_{t+1} - \theta_t \approx -\gamma [I + \rho_t H_t]^{-1} \nabla \mathcal{L}_t(\theta_t),
\end{equation}
where 
$\rho_t = \gamma (1-\alpha) (1-\beta)/\alpha \ge 0$. 
(Full derivation in Appendix~\ref{ihessian}.)
\end{proposition}

\begin{wraptable}{r}{0.5\textwidth} 
\centering
\vspace{-0.4cm}
\caption{A Unified View of CL methods as Hessian Approximation.}
\label{tab:unified_view}
\scalebox{0.85}{
\begin{tabular}{@{}lll@{}}
\toprule
Method Family & Hessian Approximation \\ \midrule
Regularization (EWC) \cite{Kirkpatrick2016OvercomingCF} & $\approx (\sum_{i=1}^{j-1} H_i)^{-1}$ \\
Meta-CL (La-MAML) \cite{lamaml} & $\approx (H_M^j)^{-1}$ \\
\textbf{SPIN} & $\approx \gamma(I + \rho_t H_t)^{-1}$ \\ 
\bottomrule
\end{tabular}
}
\vspace{0.1cm}
\end{wraptable}

Recent work~\cite{wu2023meta} unifies regularization-based methods and Meta-Continual Learning by showing that they both approximate a unified iterative update of the form:
\begin{equation} \label{eq:unified_form}
\theta \leftarrow \theta - \alpha \tilde{H}^{-1} \nabla \mathcal{L}(\theta),
\end{equation}
where $\tilde{H}^{-1}$ represents a specific (explicit or implicit) inverse Hessian approximation (\eg, $H_{reg}^{-1}$ from past-task penalties or $H_M^{-1}$ from bi-level meta-objectives).
Upon comparing our derived effective update for SPIN (Eq.~\ref{eq:sf_effective_update_final_revised}) with this unified form (Eq.~\ref{eq:unified_form}), a clear structural equivalence emerges. The SPIN optimizer, despite its distinct formulation, implicitly arrives at the same general structure by effectively using $\tilde{H}^{-1} \approx \gamma (I + \rho_t H_t)^{-1}$ as its inverse Hessian approximation. This finding allows us to surprisingly place SPIN optimization within the same conceptual framework as regularization and meta-CL methods, as summarized in ~\Cref{tab:unified_view}. This derived update rule (Eq.~\ref{eq:sf_effective_update_final_revised}) provides a new lens for understanding SPIN's underlying mechanism, revealing a deep and previously unnoted connection to a unified theoretical framework that links seemingly disparate CL approaches.

\subsection{Theoretical Analysis and Empirical Validation}
\label{sec4.3}

\begin{figure}[t]
  \centering
  \includegraphics[width=0.95\linewidth, trim=1.2cm 7cm 0.8cm 3cm, clip]{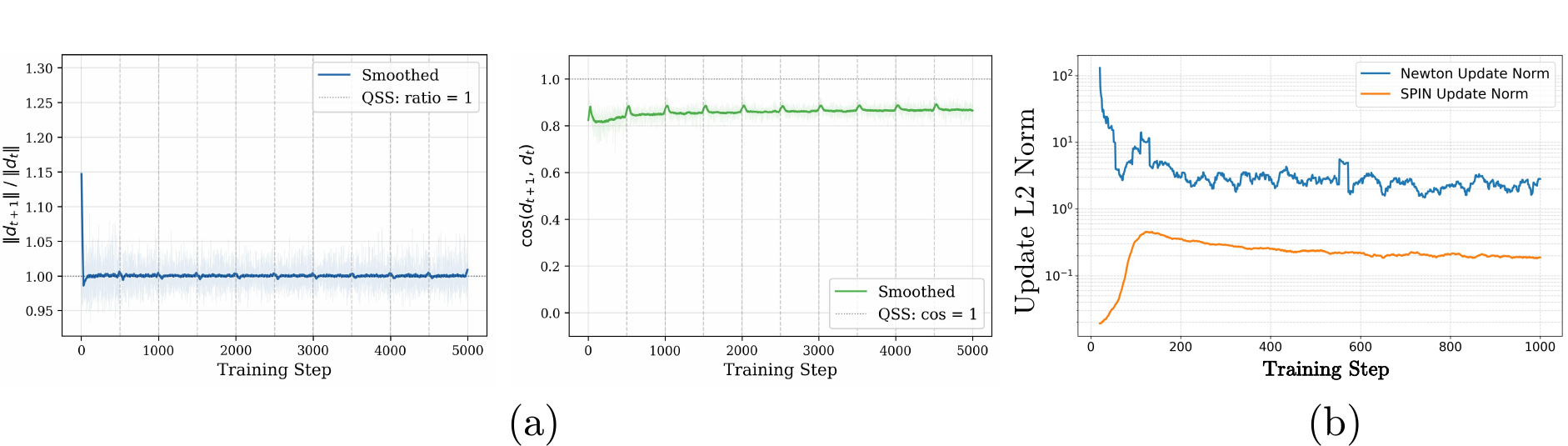}
  \caption{\textbf{Empirical validation of SPIN.} (a) Verification of the quasi-steady state assumption. (b) Stability of the update L2 norm compared to a baseline optimizer. }
  \label{fig:emp}
\end{figure}

\textbf{Theoretical Mechanisms.} Our analysis in \Cref{sec4.2} reveals that SPIN implicitly utilizes an effective inverse Hessian approximation of the form $\tilde{H}^{-1} \approx \gamma(I + \rho_t H_t)^{-1}$. A primary difficulty in OCL arises from high-variance data streams~\cite{wu2023meta}, where the local Hessian $H_t$ estimated from individual samples is extremely noisy and ill-conditioned. Conventional CL methods relying on purely data-dependent Hessian approximations risk catastrophic forgetting because they over-trust unreliable low-curvature estimates. In contrast, SPIN's formulation adds a ``floor'' (akin to Tikhonov regularization~\cite{tikhonov,tikhonov2}). This effectively damps extreme curvature ratios while preserving relative geometry. It acts as a curvature-aware method in high-curvature directions but gracefully reverts to a stable update in uncertain, low-curvature directions.

\textbf{Empirical Validation.} We empirically validate these core mechanisms. (For detailed experimental setup, see the appendix.) First, our derivation relies on the quasi-steady state (QSS) assumption ($d_{t+1} \approx d_t$). \Cref{fig:emp}a validates this assumption, showing that  $\frac{\|d_{t+1}\|}{\|d_t\|}\approx 1$ throughout training, indicating stable magnitude. Meanwhile, $\cos(d_{t+1}, d_t)$ remains close to 1, implying strong directional consistency. Notably, both properties remain stable even around task boundaries, indicating that $d_t$ evolves smoothly despite distribution shifts, which justifies the QSS assumption. Second, we validate our theory by comparing the L2 norm of the optimization updates, as this behavior is the direct, measurable symptom of the underlying scaling factor stability our theory addresses. As \Cref{fig:emp}b illustrates, the magnitude of SPIN's updates (orange line) is stable and ``not violently fluctuating,'' in stark contrast to the naive optimizer's volatile updates (blue line). 

Besides, we have the following theorem to further characterize the performance of SPIN.
\begin{theorem}[Regret Bound]\label{theorem:1}
Assume the loss  $\mathcal{L}_t$ are convex, $G$-Lipschitz and $\eta$-smooth over a closed convex domain 
$\mathcal X\subset\mathbb R^d$ with diameter $D$, and that the gradients are unbiased with bounded variance 
$\mathbb E[\|g_t-\nabla \mathcal{L}_t(x_t)\|^2]\le\sigma^2$ and bounded magnitude $\|g_t\|\le G$.  
Let $\theta^\star=\arg\min_{x\in\mathcal X}\sum_{t=1}^T \mathcal{L}_t(\theta)$ and define the static regret 
$R_T=\sum_{t=1}^T\big(\mathcal{L}_t(\theta_t)-\mathcal{L}_t(\theta^\star)\big)$.  
Let $\alpha\in(0,1)$ and step size 
$\gamma^\star=\tfrac{D}{G\sqrt{T}}\sqrt{\tfrac{\alpha}{2-\alpha}}$.  
Then with probability at least $1-\delta$ for any $\delta\in(0,1)$, we have
\begin{align}
\textstyle
R_T
\le
DG(\sqrt{\tfrac{(2-\alpha)T}{\alpha}} +\tfrac{2}{3}\ln\tfrac{1}{\delta})
+ D\sigma\sqrt{2T\ln\tfrac{1}{\delta}}.
\end{align}
\end{theorem}

\Cref{theorem:1} shows that, under mild  assumptions, our proposed optimizer achieves a nearly-optimal $\tilde{\mathcal O}(\sqrt{T})$ regret bound with high probability in the online optimization setting. This result theoretically substantiates the efficiency and stability of the proposed optimization framework. Detailed proof is provided in Appendix~\ref{rb}.

\section{Experiments}
\subsection{Experimental Setups}

\textbf{Datasets.} We evaluate our method on four datasets, \ie , CIFAR-100 (100 classes)~\cite{cifar100}, Tiny-ImageNet (200 classes)~\cite{tiny-imagenet}, CLEAR100 (100 classes)~\cite{clear100}, and Imagenet-1K (1000 classes)~\cite{imagenet1k}. More details are provided in Appendix~\ref{datasetd}.

\noindent \textbf{Baselines.} To show the effectiveness of our proposed optimizer, we integrate it into 11 representative OCL methods, which span various categories, include ER~\cite{Rolnick2018ExperienceRF}, DER++~\cite{derpp}, COPE~\cite{cope}, SCR~\cite{SCR}, ER-ACE~\cite{Caccia2021NewIO}, OCM~\cite{ocm}, OnPro~\cite{onpro}, GSA~\cite{GSA}, MOSE~\cite{yan2024orchestrate}, CCL-DC~\cite{ccldc}, S6MOD~\cite{s6mod}. To rigorously isolate and evaluate the contribution of our optimization strategy, 
we compare the performance of these OCL methods under our \textbf{SPIN} optimizer and that under four widely-used generic optimizers: SGD~\cite{SGD}, Adam~\cite{Adam}, AdamW~\cite{AdamW} and RMSProp~\cite{SGD}.

\noindent \textbf{Implementation details.} For all experiments, we employ a ResNet-18~\cite{resnet} backbone, by following the common architecture in the broad OCL literature~\cite{ccldc,s6mod,onpro}. The streaming input batch size is set to 10 and the memory buffer batch size to 64, consistent with other methods~\cite{ccldc, s6mod}. All main results use fixed hyperparameters $\alpha=\beta=0.05$. For each OCL method, we use the same learning rate and weight decay for AdamW and SPIN-AdamW, and likewise for SGD and SPIN-SGD. All other settings follow the corresponding baseline implementations. Results are averaged over 5 runs except for ImageNet-1K, which is evaluated with a single run due to computational constraints. For evaluation, we measure the model performance using Final Average Accuracy (Acc) and Averaged Anytime Accuracy (AAA). Details are provided in Appendix~\ref{td}.

\begin{table*}[t]
\centering
\caption{
    Comparison of final \textbf{Average Accuracy} (Acc. \%) using different optimizers on \textbf{Split CIFAR-100 (M=2k)} and \textbf{Split Tiny-ImageNet (M=5k)}. 
    Our \textbf{SPIN} is benchmarked against 4 standard optimizers across 11 SOTA OCL methods. The best results are highlighted using \textbf{bold} font.
}
\label{table_result_aa}
\renewcommand{\arraystretch}{1.1}
\resizebox{\textwidth}{!}{%
\begin{tabular}{l@{}l C >{\columncolor{lightgray}\bfseries}C C C C >{\columncolor{lightgray}\bfseries}D C >{\columncolor{lightgray}\bfseries}C C C C >{\columncolor{lightgray}\bfseries}D}
\toprule

& & \multicolumn{6}{c}{\textbf{Split CIFAR-100 (10 tasks, M=2k)}} & \multicolumn{6}{c}{\textbf{Split Tiny-ImageNet (20 tasks, M=5k)}} \\
\cmidrule(lr){3-8} \cmidrule(lr){9-14} 

\multicolumn{2}{l}{\textbf{Baseline}} & {SGD} & {\textbf{SPIN-SGD}} & {RMSprop} & {Adam} & {AdamW} & {\textbf{SPIN-AdamW}}  & {SGD} & {\textbf{SPIN-SGD}} & {RMSprop} & {Adam} & {AdamW} & {\textbf{SPIN-AdamW}} \\
\midrule
ER      & (2019) & 27.94 & \textbf{33.31} & 26.38 & 31.02 & 33.07 & \textbf{35.92} & 21.68 & \textbf{25.24} & 19.25 & 23.38 & 23.90 & \textbf{25.94} \\
DER++   & (2020) & 26.10 & \textbf{35.90} & 27.17 & 30.84 & 31.11 & \textbf{36.06} & 18.95 & \textbf{27.27} & 23.37 & 26.34 & 26.69 & \textbf{29.79} \\
COPE    & (2021) & 30.81 & \textbf{34.18} & 31.12 & 32.90 & 33.55 & \textbf{35.63} & 18.31 & \textbf{20.43} & 18.01 & 19.80 & 20.77 & \textbf{21.66} \\
SCR     & (2021) & 29.82 & \textbf{31.53} & 28.07 & 30.52 & 31.32 & \textbf{32.86} & 17.26 & \textbf{18.77} & 17.11 & 18.30 & 18.74 & \textbf{20.39} \\
ER-ACE  & (2022) & 28.54 & \textbf{36.14} & 27.65 & 32.43 & 34.77 & \textbf{37.02} & 24.95 & \textbf{28.11} & 23.75 & 28.10 & 28.52 & \textbf{30.42} \\
OCM     & (2022) & 33.21 & \textbf{36.06} & 33.69 & 36.78 & 37.45 & \textbf{40.15} & 22.19 & \textbf{24.91} & 23.47 & 24.83 & 25.68 & \textbf{27.29} \\
OnPro   & (2023) & 32.44 & \textbf{34.35} & 29.56 & 33.56 & 34.25 & \textbf{36.61} & 21.38 & \textbf{23.42} & 20.21 & 21.99 & 22.26 & \textbf{24.94} \\
GSA     & (2023) & 33.25 & \textbf{36.79} & 30.78 & 35.90 & 36.53 & \textbf{37.73} & 20.80 & \textbf{22.44} & 20.84 & 22.16 & 25.67 & \textbf{26.63} \\
MOSE    & (2024) & 44.18 & \textbf{47.18} & 43.46 & 46.21 & 46.95 & \textbf{48.33} & 28.21 & \textbf{29.91} & 28.02 & 29.57 & 29.87 & \textbf{30.63} \\
CCL-DC  & (2024) & 38.80 & \textbf{41.74} & 42.17 & 42.97 & 43.99 & \textbf{45.68} & 27.01 & \textbf{28.91} & 27.84 & 28.10 & 28.35 & \textbf{29.32} \\
S6MOD   & (2025) & 34.12 & \textbf{34.81} & 33.08 & 35.19 & 35.55 & \textbf{37.37} & 22.02 & \textbf{24.00} & 20.94 & 23.43 & 24.01 & \textbf{25.17} \\
\midrule
\multicolumn{2}{l}{\textbf{Average}} & 32.66 & \textbf{36.54} & 32.10 & 35.30 & 36.23 & \textbf{38.49} & 21.25 & \textbf{23.95} & 21.01 & 24.09 & 24.95 & \textbf{26.56} \\
\bottomrule
\end{tabular}} 
\end{table*}

\begin{table*}[t]
\centering
\caption{
    Comparison of \textbf{Averaged Anytime Accuracy} (AAA \%) using different optimizers on \textbf{Split CIFAR-100 (M=2k)} and \textbf{Split Tiny-ImageNet (M=5k)}. 
    Our \textbf{SPIN} is benchmarked against 4 standard optimizers across 11 SOTA OCL methods. The best results are highlighted using \textbf{bold} font.
}
\label{table_result_aaa}
\renewcommand{\arraystretch}{1.1}
\resizebox{\textwidth}{!}{%
\begin{tabular}{l@{}l C >{\columncolor{lightgray}\bfseries}C C C C >{\columncolor{lightgray}\bfseries}D C >{\columncolor{lightgray}\bfseries}C C C C >{\columncolor{lightgray}\bfseries}D}
\toprule
& & \multicolumn{6}{c}{\textbf{Split CIFAR-100 (10 tasks, M=2k)}} & \multicolumn{6}{c}{\textbf{Split Tiny-ImageNet (20 tasks, M=5k)}} \\
\cmidrule(lr){3-8} \cmidrule(lr){9-14} 

\multicolumn{2}{l}{\textbf{Baseline}} & {SGD} & {\textbf{SPIN-SGD}} & {RMSprop} & {Adam} & {AdamW} & {\textbf{SPIN-AdamW}}  & {SGD} & {\textbf{SPIN-SGD}} & {RMSprop} & {Adam} & {AdamW} & {\textbf{SPIN-AdamW}} \\
\midrule
ER      & (2019) & 36.86 & \textbf{46.18} & 34.54 & 40.68 & 41.50 & \textbf{48.08} & 28.36 & \textbf{34.38} & 28.14 & 32.47 & 32.81 & \textbf{37.33} \\
DER++   & (2020) & 35.16 & \textbf{45.49} & 33.68 & 37.99 & 38.86 & \textbf{46.26} & 21.20 & \textbf{35.39} & 30.50 & 36.49 & 36.87 & \textbf{40.08} \\
COPE    & (2021) & 37.76 & \textbf{44.62} & 34.09 & 39.48 & 42.01 & \textbf{47.38} & 26.49 & \textbf{26.80} & 24.44 & 28.54 & 27.99 & \textbf{31.01} \\
SCR     & (2021) & 42.55 & \textbf{45.70} & 37.74 & 41.24 & 42.44 & \textbf{48.20} & 24.03 & \textbf{26.01} & 23.15 & 26.19 & 26.24 & \textbf{28.30} \\
ER-ACE  & (2022) & 38.38 & \textbf{45.91} & 38.48 & 41.26 & 42.05 & \textbf{46.61} & 32.20 & \textbf{37.25} & 31.56 & 36.55 & 37.54 & \textbf{40.03} \\
OCM     & (2022) & 40.17 & \textbf{41.83} & 39.98 & 42.26 & 43.39 & \textbf{47.06} & 27.28 & \textbf{30.73} & 26.44 & 30.08 & 31.50 & \textbf{34.21} \\
OnPro   & (2023) & 38.09 & \textbf{40.29} & 37.20 & 41.21 & 41.44 & \textbf{43.97} & 22.44 & \textbf{29.86} & 24.87 & 29.41 & 28.24 & \textbf{32.48} \\
GSA     & (2023) & 44.91 & \textbf{50.43} & 39.44 & 44.40 & 45.40 & \textbf{51.37} & 27.45 & \textbf{31.16} & 27.08 & 31.87 & 33.08 & \textbf{35.93} \\
MOSE    & (2024) & 56.53 & \textbf{58.55} & 53.49 & 58.94 & 58.76 & \textbf{61.79} & 33.65 & \textbf{36.89} & 33.06 & 36.48 & 37.01 & \textbf{40.74} \\
CCL-DC  & (2024) & 47.80 & \textbf{51.29} & 49.53 & 55.30 & 55.61 & \textbf{58.58} & 34.21 & \textbf{36.52} & 33.79 & 35.13 & 35.84 & \textbf{37.48} \\
S6MOD   & (2025) & 42.21 & \textbf{45.02} & 41.35 & 44.17 & 44.67 & \textbf{47.96} & 28.08 & \textbf{29.93} & 29.24 & 30.21 & 31.00 & \textbf{32.85} \\
\midrule
\multicolumn{2}{l}{\textbf{Average}} & 41.86 & \textbf{46.85} & 39.96 & 44.27 & 45.10 & \textbf{49.75} & 27.76 & \textbf{32.27} & 28.54 & 32.13 & 32.55 & \textbf{34.59} \\
\bottomrule
\end{tabular}
} 
\end{table*}

\subsection{Main Results} 

\begin{table*}[t]
  \centering

  \begin{minipage}[t]{0.40\textwidth}
    \centering
    \captionof{table}{Comparison of final Average Accuracy (Acc.) and Averaged Anytime Accuracy (AAA) on Imagenet-1K.}
    \label{tab:1k}

    \scriptsize
    \setlength{\tabcolsep}{8.7pt}
    \renewcommand{\arraystretch}{0.65}

    \resizebox{\linewidth}{!}{
      \begin{tabular}{l|cc|cc}
        \toprule
        & \multicolumn{2}{c}{Acc.(\%)} & \multicolumn{2}{c}{AAA(\%)} \\
        \cmidrule(lr){2-3} \cmidrule(lr){4-5}
        Baseline & SGD & SPIN & SGD & SPIN \\
        \midrule
        ER      & 13.21 & \textbf{15.37} & 19.76 & \textbf{21.58} \\
        DER++   & 14.19 & \textbf{17.51} & 18.50 & \textbf{22.02} \\
        \bottomrule
      \end{tabular}
    }
  \end{minipage}
  \hfill
  \begin{minipage}[t]{0.56\textwidth}
    \centering
    \captionof{table}{Comparison on CLEAR100 dataset (Acc \%).}
    \label{table_clear100}

    \scriptsize
    \setlength{\tabcolsep}{3pt}
    \renewcommand{\arraystretch}{0.65}

    \resizebox{\linewidth}{!}{
      \begin{tabular}{l|ccc|ccc}
        \toprule
        & \multicolumn{3}{c}{M=200} & \multicolumn{3}{c}{M=5000} \\
        \cmidrule(lr){2-4} \cmidrule(lr){5-7}
        Baseline & SGD & SGD-M & SPIN & SGD & SGD-M & SPIN \\
        \midrule
        ER      & 32.63 & 36.01 & \textbf{37.99} & 41.38 & 51.87 & \textbf{53.02} \\
        DER++   & 33.54 & 36.87 & \textbf{37.96} & 41.20 & 50.49 & \textbf{52.21} \\
        GSA     & 37.76 & 40.62 & \textbf{42.38} & 43.86 & 46.20 & \textbf{47.88} \\
        \bottomrule
      \end{tabular}
    }
  \end{minipage}

\end{table*}
\textbf{Performance on Standard Benchmarks.} As shown in \Cref{table_result_aa} and \Cref{table_result_aaa}, SPIN consistently outperforms generic optimizers across 11 OCL baselines on Split CIFAR-100 and Tiny-ImageNet. 
Notably, SPIN-AdamW raises MOSE's final accuracy (ACC) from 46.95\% to 48.33\% and GSA's anytime accuracy (AAA) from 45.40\% to 51.37\% on CIFAR-100. This demonstrates its strong ability to improve peak performance and effectively mitigate forgetting.

\textbf{Scalability to Million-Scale Data and Robustness to Natural Distribution Shifts.} On the large-scale ImageNet-1K dataset, replacing SGD with SPIN yields substantial gains (\Cref{tab:1k}). For instance, it improves DER++'s ACC from 14.19\% to 17.51\% and AAA from 18.50\% to 22.02\%, confirming its effectiveness and scalability on complex data streams. On CLEAR100, which exhibits real world natural shifts devoid of clear task boundaries, SPIN consistently surpasses both standard SGD and SGD-M, i.e., SGD based on Exponential Moving Average (EMA), across varying memory sizes (\Cref{table_clear100}), proving its stability-plasticity coupling is superior to naive parameter averaging.

\subsection{Ablation Studies and Analysis}
\label{ablation}

To further understand the behavior of SPIN, we conduct a series of ablation studies and analyses to answer the following key questions:

\begin{wrapfigure}{r}{0.56\textwidth}
    \centering
    \vspace{-1em}
    
    \includegraphics[width=\linewidth]{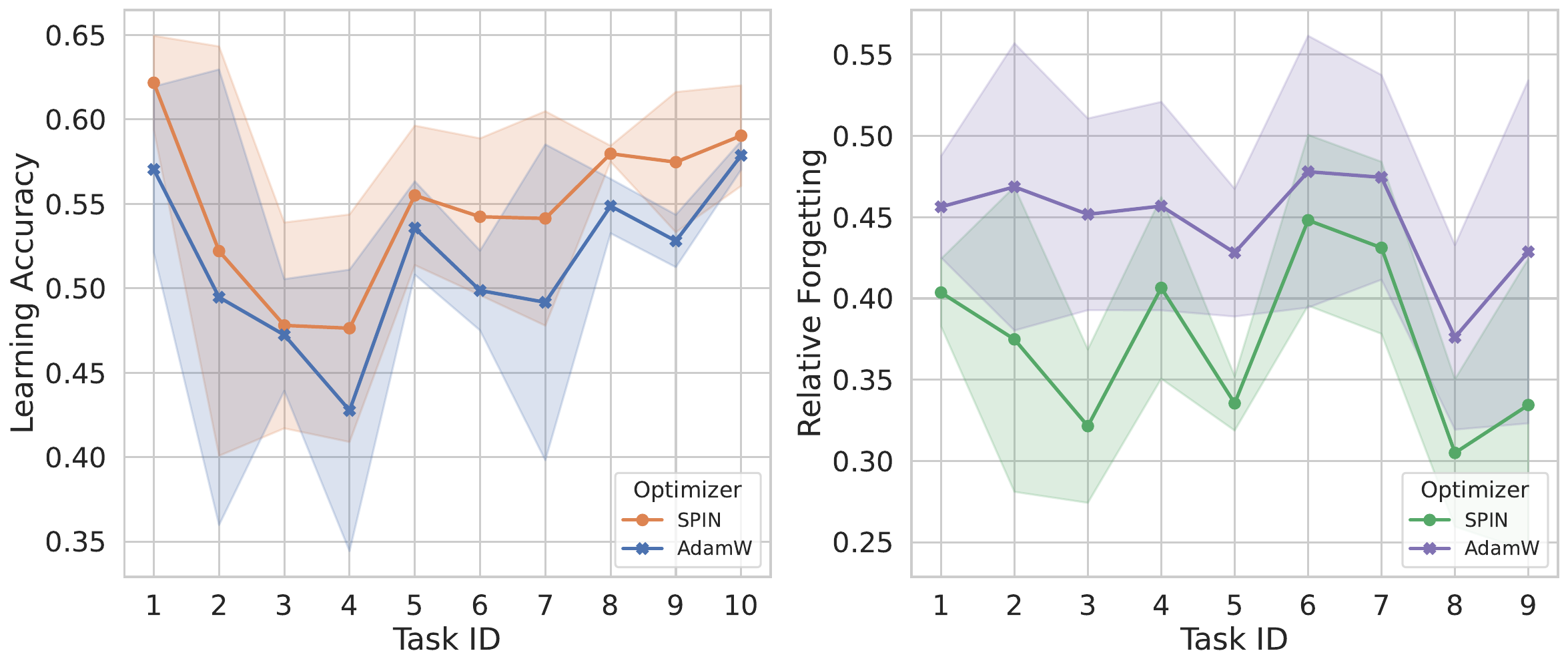}
    
    \caption{Task-wise Learning Accuracy and Relative Forgetting of the ER method ($M=2000$), comparing the performance of AdamW and SPIN.}
    \label{fig5}
    
    \vspace{-1em}
\end{wrapfigure}

\textbf{Q1: Does SPIN effectively balance stability and plasticity?}

To better understand the learning dynamics,~\Cref{fig5} compares the task-wise Learning Accuracy (reflecting plasticity) 
and Relative Forgetting (reflecting stability) between SPIN and AdamW. SPIN consistently achieves higher learning accuracy 
while maintaining lower relative forgetting throughout the task stream. These results suggest that SPIN enables effective 
adaptation to new tasks without excessively overwriting previously acquired knowledge. This behavior is consistent with 
the core design of SPIN, where the interpolation between plastic and stable iterate sequences promotes a better balance 
between adaptation and retention during continual optimization. \newtext{We further compare Relative Forgetting(RF) across 11 OCL methods on Split CIFAR-100 (\Cref{tab:relative_forgetting}). SPIN lowers average 
RF from 31.94\% to 29.99\% with SGD and from 30.83\% to 28.63\% with AdamW. Combined with the higher final accuracies in 
\Cref{table_result_aa}, these results indicate a better overall accuracy--forgetting trade-off.}

\begin{table}[!htbp]
\centering
\captionsetup{position=top,skip=7pt}
\footnotesize
\renewcommand{\arraystretch}{0.85}
\setlength{\tabcolsep}{3pt}
\sbox{\rfTableBox}{\begin{tabular}{lrrrr}
\toprule
Method & SGD & SPIN-SGD & AdamW & SPIN-AdamW \\
\midrule
ER      & 39.14 & \textbf{36.70} & 36.15 & \textbf{33.09} \\
DER++   & 53.30 & \textbf{42.41} & 45.43 & \textbf{36.49} \\
COPE    & 13.64 & \textbf{12.50} & 12.07 & \textbf{10.91} \\
SCR     & \textbf{50.47} & 51.33 & \textbf{46.98} & 50.19 \\
ER-ACE  & \textbf{17.24} & 19.27 & 17.20 & \textbf{16.72} \\
OCM     & 14.01 & \textbf{13.78} & 13.51 & \textbf{12.98} \\
OnPro   & 16.40 & \textbf{15.24} & 21.88 & \textbf{17.46} \\
GSA     & \textbf{36.53} & 39.45 & 35.76 & \textbf{34.76} \\
MOSE    & 33.47 & \textbf{31.35} & 37.20 & \textbf{36.11} \\
CCL-DC  & 37.56 & \textbf{33.78} & 37.62 & \textbf{32.05} \\
S6MOD   & 39.62 & \textbf{34.11} & 35.35 & \textbf{34.20} \\
\midrule
Average & 31.94 & \textbf{29.99} & 30.83 & \textbf{28.63} \\
\bottomrule
\end{tabular}}
\newcommand{\ablationtabular}{\begin{tabular}{l c c}
      \toprule
      Method & Acc.(\%) $\uparrow$ & AAA(\%) $\uparrow$  \\
      \midrule
      AdamW & 27.84 & 39.53 \\
      SPIN($\alpha$=0) & 1.04 & 2.93 \\
      SPIN($\alpha$=1) & 27.52 & 37.70 \\
      SPIN($\beta$=1) & 22.30 & 30.63 \\
      SPIN($\beta$=0) & 31.34 & 40.28 \\
      SPIN & \textbf{32.20} & \textbf{42.39} \\
      \bottomrule
    \end{tabular}}
\begin{minipage}[t]{0.56\textwidth}
\vspace{0pt}
\caption{Relative Forgetting(RF) (\% $\downarrow$) on Split CIFAR-100 ($M=2$k). Best in each optimizer pair is bolded.}
\label{tab:relative_forgetting}
\centering
\usebox{\rfTableBox}
\end{minipage}\hfill
\begin{minipage}[t]{0.42\textwidth}
\vspace{0pt}
\caption{Component ablation of SPIN-AdamW with GSA on Split CIFAR-100 ($M=1$k), comparing boundary settings of $\alpha$ and $\beta$. ACC and AAA (\%) are reported. Best results are bolded.}
\label{tab:ablation}
\centering
\setlength{\tabcolsep}{5pt}
\renewcommand{\arraystretch}{1.15}
\ablationtabular
\end{minipage}
\end{table}

\textbf{Q2: What are the key components behind SPIN’s effectiveness?} 
We first analyze several boundary cases of SPIN in ~\Cref{tab:ablation}. 
When $\alpha=0$, the optimization collapses since the model parameters remain nearly unchanged throughout training. 
When $\alpha=1$, SPIN degenerates into standard AdamW. 
Similarly, when $\beta=1$, gradients are evaluated entirely on the stability sequence, resulting in overly conservative updates that hinder adaptation to new tasks. 
When $\beta=0$, SPIN reduces to an EMA-based optimizer (\ie, AdamW+EMA), similar to prior temporal ensemble approaches~\cite{soutifcormerais2023improving}. 
Although EMA improves optimization stability, SPIN consistently achieves superior performance across all settings. 
Additional comparisons with SGD+EMA baselines are provided in  Appendix~\ref{hpema}. 
These results suggest that the gains of SPIN cannot be solely attributed to temporal averaging, and instead highlight the importance of gradient-location interpolation between the plastic and stable sequences.

\begin{figure*}[t]
\centering

\begin{minipage}[t]{0.45\textwidth}
\vspace{0pt}
\centering

\captionof{table}{
Comparison of \textbf{Averaged Anytime Accuracy} (AAA \%) and 
\textbf{Average Accuracy} (ACC \%) using different Rehearsal-Free 
methods on \textbf{Split CIFAR-10} and \textbf{Split CIFAR-100}. 
The best and second best results are highlighted using 
\textbf{bold} and \underline{underline} fonts.
}
\label{ref_acc}

\resizebox{\linewidth}{!}{%
\begin{tabular}{l c c c c}
\toprule
& \multicolumn{2}{c}{\textbf{Split CIFAR-10}} 
& \multicolumn{2}{c}{\textbf{Split CIFAR-100}} \\
\cmidrule(lr){2-3} 
\cmidrule(lr){4-5}

\textbf{Baseline} & AAA & Acc & AAA & Acc \\
\midrule

on-EWC      & 40.27 & 18.46 & 13.35 & 4.99 \\
SI       & 39.84 & 17.49 & 12.32 & 5.20 \\
SPIN-SGD    & \underline{40.52} & \underline{19.09} 
             & \underline{14.63} & \textbf{5.91} \\
SPIN-AdamW  & \textbf{41.16} & \textbf{19.18} 
             & \textbf{15.60} & \underline{5.58} \\

\bottomrule
\end{tabular}
}
\end{minipage}
\hfill
\begin{minipage}[t]{0.44\textwidth}
\vspace{0pt}
\centering

\includegraphics[width=0.47\linewidth]
{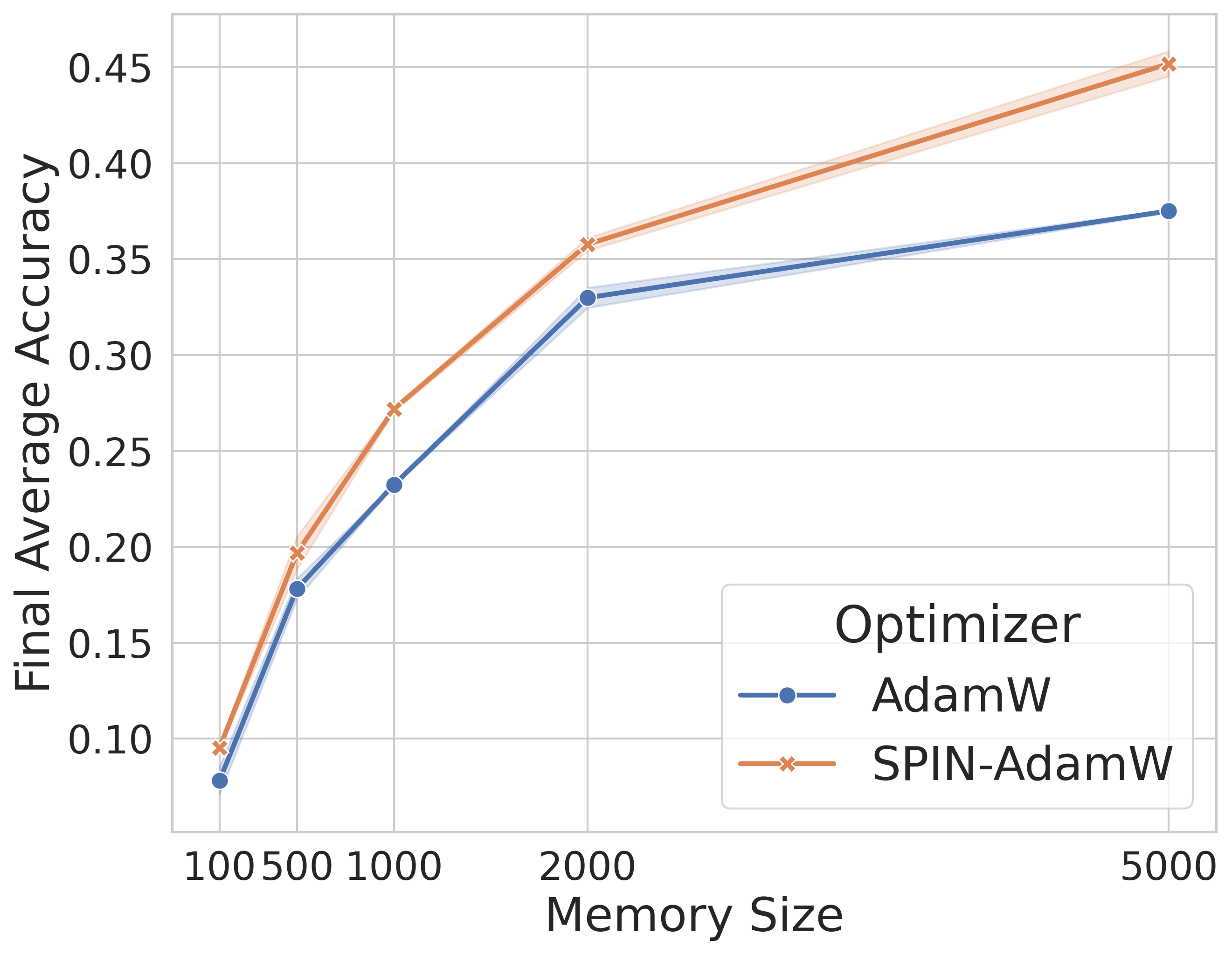}
\hfill
\includegraphics[width=0.47\linewidth]
{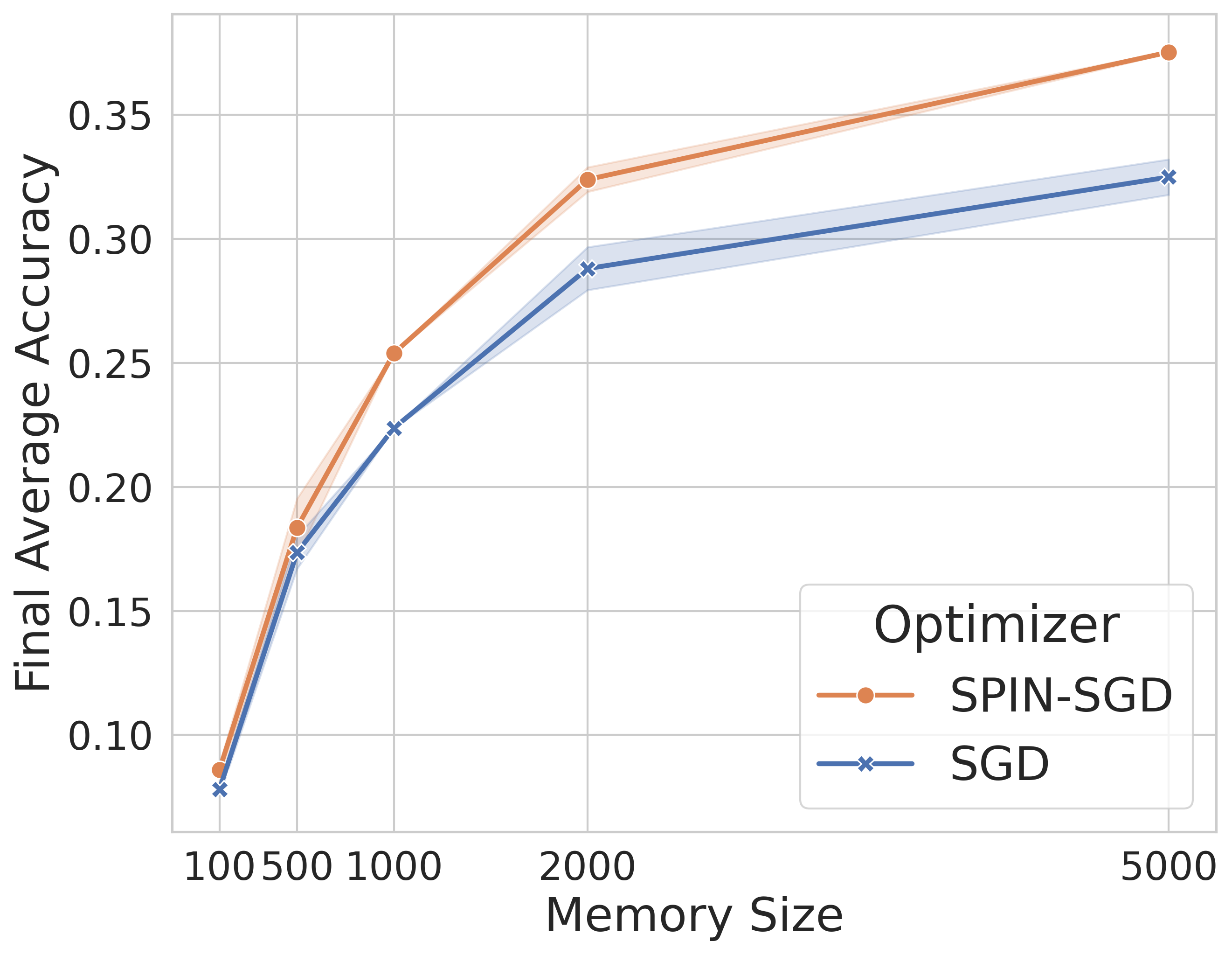}

\vspace{0.2cm}

\includegraphics[width=0.47\linewidth]
{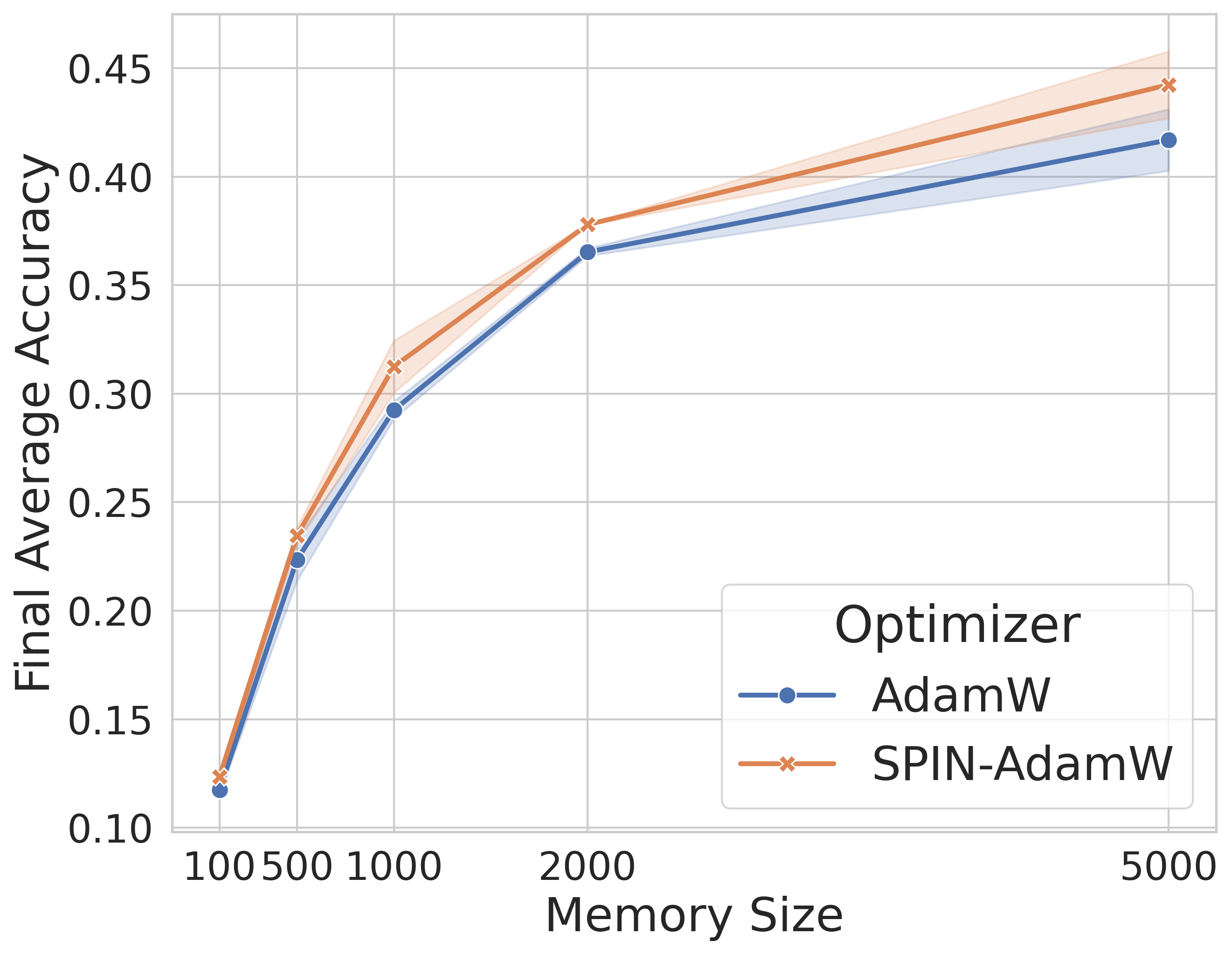}
\hfill
\includegraphics[width=0.47\linewidth]
{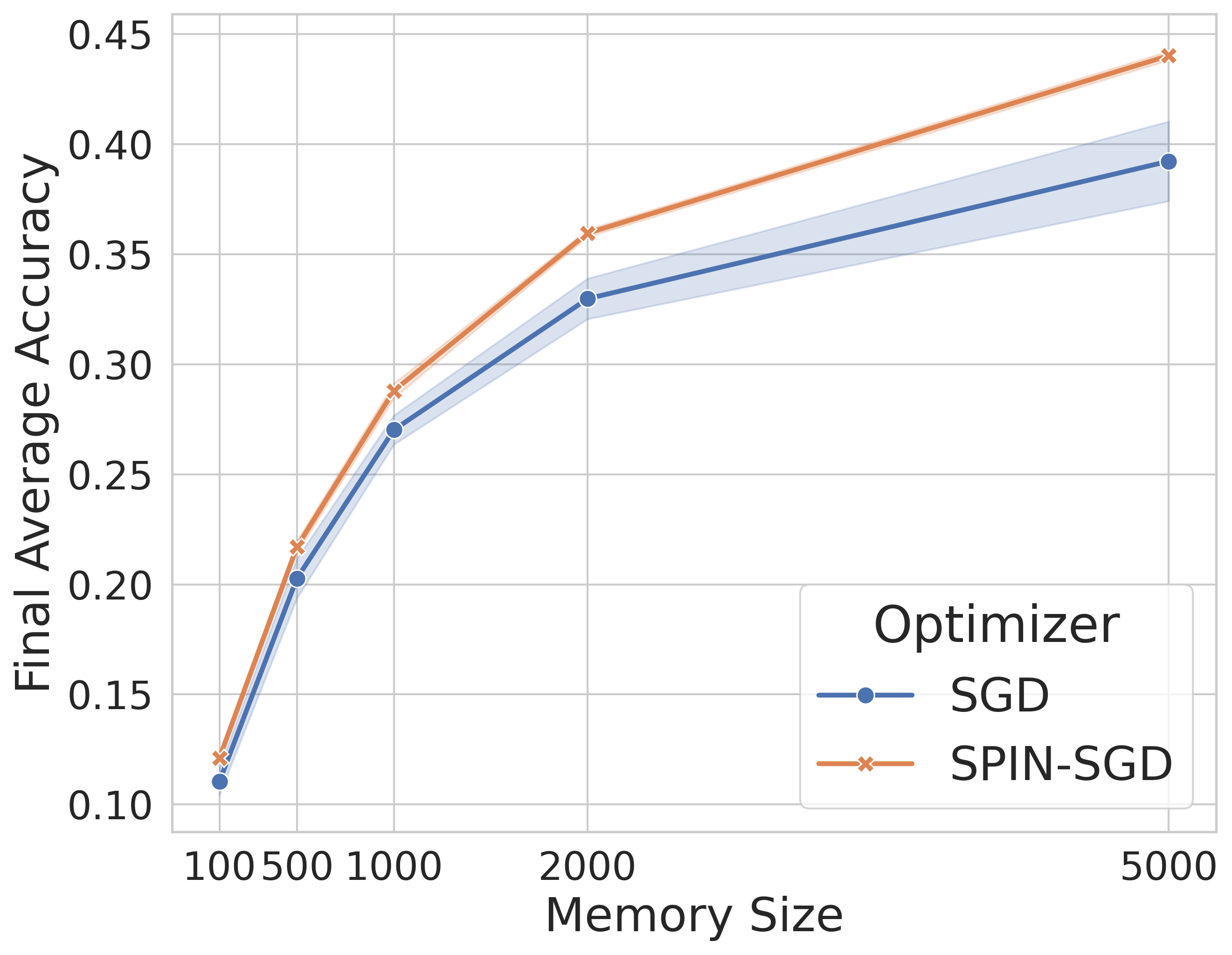}

\captionof{figure}{Impact of Replay Buffer Size}
\label{fig:mem_size_accuracy}

\end{minipage}

\end{figure*}

\textbf{Q3: Is SPIN robust under different memory settings?} 
An ideal CL method should be robust across different memory sizes in real world application. To investigate whether SPIN is effective with different memory constraints, we conducted experiments on CIFAR-100 using various memory sizes. As seen from \Cref{fig:mem_size_accuracy}, SPIN consistently improves performance across all replay buffer sizes, including strict memory-constrained settings, suggesting that SPIN is effective even with different buffer sizes.

\textbf{Q4: Can SPIN maintain stability without replay memory?} 
To further investigate whether SPIN can induce intrinsic stability beyond replay mechanisms, we evaluate SPIN under the rehearsal-free setting. 
We compare SPIN against representative rehearsal-free continual learning methods, including On-EWC~\cite{Huszr2017NoteOT} and SI~\cite{SI}, which explicitly regularize parameter updates using importance-aware constraints.
As shown in~\Cref{ref_acc}, both SPIN-SGD and SPIN-AdamW consistently outperform these rehearsal-free baselines across all metrics. 
These results suggest that SPIN is able to preserve past knowledge through its optimization dynamics alone, which is consistent with our theoretical analysis that SPIN implicitly induces Hessian-aware stabilization during continual optimization.

\begin{table}[t]
\centering
\captionsetup{position=top,skip=7pt}
\footnotesize
\renewcommand{\arraystretch}{1.285}
\sbox{\runtimeTableBox}{\resizebox{0.485\textwidth}{!}{\begin{tabular}{l|cc}
            \toprule
            \textbf{Method} & \textbf{Runtime}  & \textbf{Peak Memory} \\
            \midrule
            SGD & 1.04(ms/step) & 155.86(MB) \\
            SGD+EMA & 1.08(ms/step) & 240.75(MB) \\
            AdamW & 1.68(ms/step) & 267.58(MB) \\
            \rowcolor{gray!15} \textbf{SPIN-SGD} & \textbf{1.13}(ms/step) & \textbf{240.75}(MB) \\
            \bottomrule
        \end{tabular}}}
\renewcommand{\arraystretch}{1.198}
\sbox{\memoryTableBox}{\fontsize{9}{9.5}\selectfont\begin{tabular*}{0.485\textwidth}{@{\extracolsep{\fill}}l|ccc}
            \toprule
            & \multicolumn{2}{c}{\textbf{SGD}} & \textbf{SPIN} \\
            \cmidrule(lr){2-3} \cmidrule(lr){4-4}
            \textbf{Buffer Size} & \textbf{15k} & \textbf{30k} & \textbf{15k}\\
            \midrule
            \textbf{Acc (\%)} & 41.61 & 46.93 & \textbf{48.46} \\
            \textbf{AAA (\%)} & 47.41 & 51.97 & \textbf{55.67} \\
            \bottomrule
        \end{tabular*}}
\begin{minipage}[t]{0.485\textwidth}
\vspace{0pt}
\caption{Runtime per training step (ms/step) and peak memory (MB) of different optimizers with GSA on split CIFAR-100.}
\label{tab:runtime}
\centering
\usebox{\runtimeTableBox}
\end{minipage}\hfill%
\begin{minipage}[t]{0.485\textwidth}
\vspace{0pt}
\caption{Accuracy under approximately matched total memory with GSA on CIFAR-100. SGD uses a 30k replay buffer to offset SPIN's additional optimizer states with a 15k buffer. Best results are bolded.}
\label{tab:rebuttal_exp}
\centering
\usebox{\memoryTableBox}
\end{minipage}
\end{table}

\textbf{Q5: Are SPIN’s gains simply due to increased optimizer memory?} 
As shown in~\Cref{tab:runtime}, SPIN-SGD introduces only modest runtime and memory overhead compared to vanilla SGD, while remaining substantially more efficient than AdamW. 
To rule out the possibility that SPIN’s gains arise merely from additional optimizer states, we evaluate SPIN under a strict equal-memory budget. 
On CIFAR-100, the extra optimizer state of SPIN-SGD is equivalent to approximately 15k replay samples. 
We therefore compare SPIN with a 15k replay buffer against SGD with a 30k replay buffer, ensuring comparable total memory usage. 
As shown in~\Cref{tab:rebuttal_exp}, SPIN (15k) still achieves higher ACC and AAA than the memory-matched SGD (30k). 
These results suggest that SPIN’s improvements primarily arise from its optimization dynamics and stability–plasticity balancing mechanism.

\begin{figure}[t]
  \centering
  \begin{subfigure}[t]{0.55\linewidth}
    \centering
    \includegraphics[width=\linewidth]{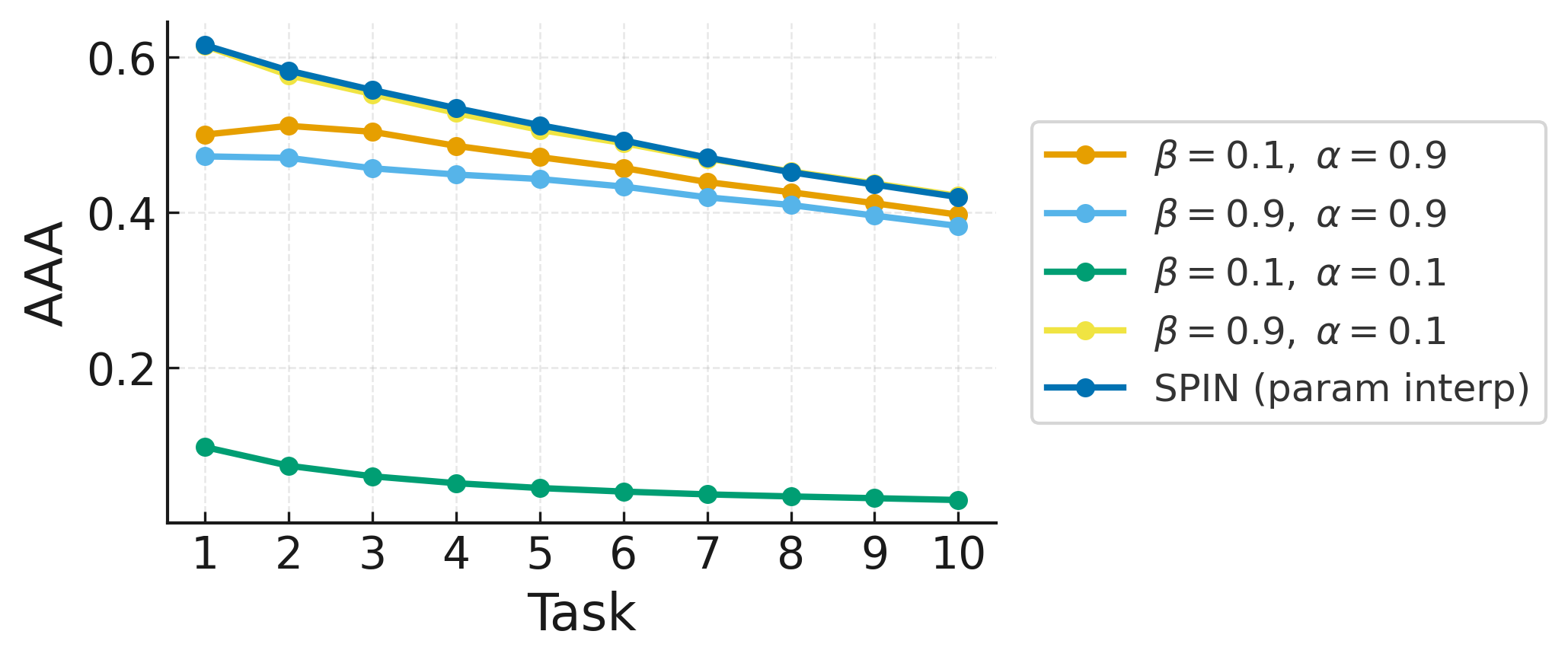}
    \caption{Task-wise AAA under different averaging mechanisms.}
    \label{fig:sub1}
  \end{subfigure}
  \begin{subfigure}[t]{0.36\linewidth}
    \centering
    \includegraphics[width=\linewidth]{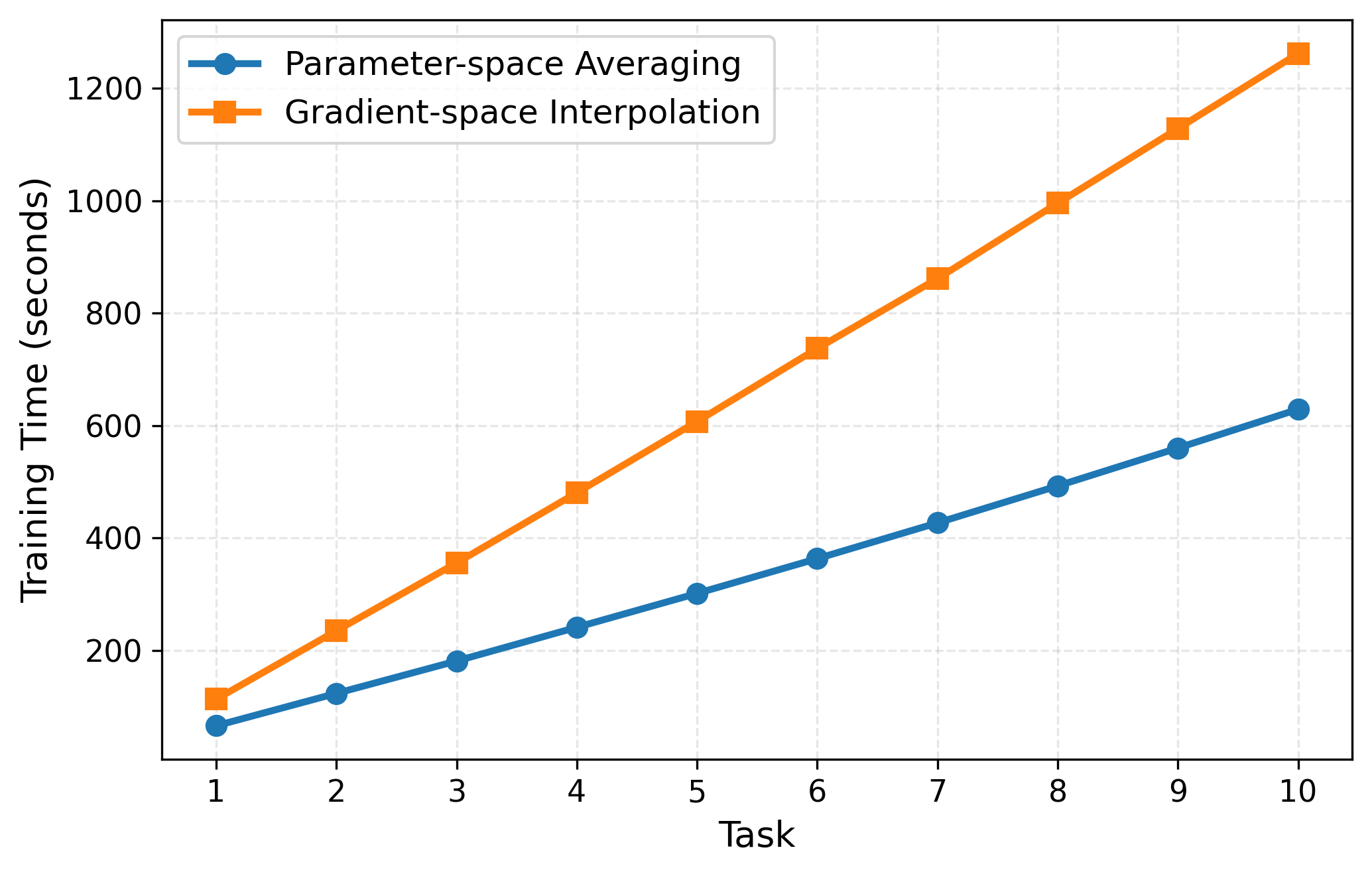}
    \caption{Training time comparison.}
    \label{fig:sub2}
  \end{subfigure}
  \caption{Comparison between parameter-space interpolation and gradient-space averaging in OCL.}
  \label{fig:ev}

\end{figure}

\textbf{Q6: Which averaging mechanism is more suitable for OCL?}
We compare parameter-space interpolation with gradient-space averaging under the same optimization framework. As shown in ~\Cref{fig:sub1}, parameter-space interpolation consistently achieves higher task-wise AAA, suggesting that evaluating gradients at a shared interpolated point yields a more coherent optimization trajectory under non-stationary data streams. In contrast, directly averaging gradients computed at different parameter locations introduces conflicting update directions.
We further compare their computational efficiency in ~\Cref{fig:sub2}. Since gradient-space averaging requires two forward-backward passes, its training time is nearly twice that of parameter-space interpolation. In contrast, SPIN only requires a single gradient evaluation while achieving better continual learning performance.

\section{Conclusion and Limitations}

In this paper, we investigate the optimizer design for OCL and introduce a novel and dedicated optimizer SPIN, which explicitly balances the stability and plasticity tradeoff. Through its interpolation-based update mechanism, SPIN achieves smooth and robust optimization under non-stationary data streams, effectively mitigating catastrophic forgetting. Extensive experiments across diverse OCL frameworks demonstrate that SPIN consistently improves both accuracy and stability compared to widely used generic optimizers, regardless of the presence or size of replay buffers and even under strict memory constraints. These findings highlight that optimizer design plays a critical role in achieving a stable learning–forgetting balance.  

\textbf{Limitations:} Our theoretical analysis relies on smoothness assumptions and the quasi-steady-state approximation, which may not fully characterize the optimization dynamics of highly over-parameterized large-scale neural networks. In addition, although maintaining dual iterate sequences brings substantial continual learning improvements, it also introduces additional optimizer states compared to standard optimizers. Exploring more parameter-efficient interpolation mechanisms that maintain the trade-off between stability and plasticity, thereby making them more effective on edge devices, remains an important direction for future research. \newtext{Another promising direction is to integrate SPIN with continual learning methods built on large-scale pretrained models, such as CLIP and pretrained ViTs, and investigate its role in balancing adaptation and retention during prompt tuning and low-rank adaptation.}

\begin{ack}
\newtext{\textbf{Acknowledgments.} We thank the anonymous reviewers and the area chair for their constructive feedback, which helped improve this work.}

\newtext{\textbf{Funding.} The authors received no external funding or
third-party support for this work.}

\newtext{\textbf{Competing interests.} The authors declare no competing interests.}
\end{ack}

\bibliographystyle{plainnat}
\bibliography{references}

\begin{thebibliography}{77}
\providecommand{\natexlab}[1]{#1}
\providecommand{\url}[1]{\texttt{#1}}
\expandafter\ifx\csname urlstyle\endcsname\relax
  \providecommand{\doi}[1]{doi: #1}\else
  \providecommand{\doi}{doi: \begingroup \urlstyle{rm}\Url}\fi

\bibitem[Ahn et~al.(2025)Ahn, Magakyan, and Cutkosky]{ahn2025general}
Kwangjun Ahn, Gagik Magakyan, and Ashok Cutkosky.
\newblock General framework for online-to-nonconvex conversion: Schedule-free {SGD} is also effective for nonconvex optimization.
\newblock In \emph{Proceedings of the 42nd International Conference on Machine Learning}, volume 267 of \emph{Proceedings of Machine Learning Research}, pages 772--795. PMLR, 2025.
\newblock URL \url{https://proceedings.mlr.press/v267/ahn25a.html}.

\bibitem[Arani et~al.(2022)Arani, Sarfraz, and Zonooz]{arani2022learning}
Elahe Arani, Fahad Sarfraz, and Bahram Zonooz.
\newblock Learning fast, learning slow: A general continual learning method based on complementary learning system.
\newblock In \emph{International Conference on Learning Representations}, 2022.
\newblock URL \url{https://openreview.net/forum?id=uxxFrDwrE7Y}.

\bibitem[Arbel and Mairal(2021)]{qss2}
Michal Arbel and Julien Mairal.
\newblock Amortized implicit differentiation for stochastic bilevel optimization.
\newblock \emph{ArXiv}, abs/2111.14580, 2021.
\newblock URL \url{https://api.semanticscholar.org/CorpusID:244714983}.

\bibitem[Attouch et~al.(2023)Attouch, Balhag, Chbani, and Riahi]{tikhonov2}
Hedy Attouch, A{\"\i}cha Balhag, Zaki Chbani, and Hassan Riahi.
\newblock Accelerated gradient methods combining tikhonov regularization with geometric damping driven by the hessian.
\newblock \emph{Applied Mathematics \& Optimization}, 88\penalty0 (2):\penalty0 29, 2023.

\bibitem[Bo{\c{t}} et~al.(2021)Bo{\c{t}}, Csetnek, and L{\'a}szl{\'o}]{tikhonov}
Radu~Ioan Bo{\c{t}}, Ern{\"o}~Robert Csetnek, and Szil{\'a}rd~Csaba L{\'a}szl{\'o}.
\newblock Tikhonov regularization of a second order dynamical system with hessian driven damping.
\newblock \emph{Mathematical Programming}, 189\penalty0 (1):\penalty0 151--186, 2021.

\bibitem[Buzzega et~al.(2020)Buzzega, Boschini, Porrello, Abati, and Calderara]{derpp}
Pietro Buzzega, Matteo Boschini, Angelo Porrello, Davide Abati, and Simone Calderara.
\newblock Dark experience for general continual learning: a strong, simple baseline.
\newblock \emph{ArXiv}, abs/2004.07211, 2020.
\newblock URL \url{https://api.semanticscholar.org/CorpusID:215768806}.

\bibitem[Caccia et~al.(2021)Caccia, Aljundi, Asadi, Tuytelaars, Pineau, and Belilovsky]{Caccia2021NewIO}
Lucas Caccia, Rahaf Aljundi, Nader Asadi, Tinne Tuytelaars, Joelle Pineau, and Eugene Belilovsky.
\newblock New insights on reducing abrupt representation change in online continual learning.
\newblock \emph{ArXiv}, abs/2203.03798, 2021.
\newblock URL \url{https://api.semanticscholar.org/CorpusID:247315339}.

\bibitem[Chaudhry et~al.(2018)Chaudhry, Ranzato, Rohrbach, and Elhoseiny]{agem}
Arslan Chaudhry, Marc'Aurelio Ranzato, Marcus Rohrbach, and Mohamed Elhoseiny.
\newblock Efficient lifelong learning with a-gem.
\newblock \emph{ArXiv}, abs/1812.00420, 2018.
\newblock URL \url{https://api.semanticscholar.org/CorpusID:54443381}.

\bibitem[De~Lange et~al.(2022)De~Lange, Aljundi, Masana, Parisot, Jia, Leonardis, Slabaugh, and Tuytelaars]{DeSurvey}
Matthias De~Lange, Rahaf Aljundi, Marc Masana, Sarah Parisot, Xu~Jia, Aleš Leonardis, Gregory Slabaugh, and Tinne Tuytelaars.
\newblock A continual learning survey: Defying forgetting in classification tasks.
\newblock \emph{IEEE Transactions on Pattern Analysis and Machine Intelligence}, 44\penalty0 (7):\penalty0 3366--3385, 2022.
\newblock \doi{10.1109/TPAMI.2021.3057446}.

\bibitem[Defazio(2020)]{primal2}
Aaron Defazio.
\newblock Momentum via primal averaging: Theoretical insights and learning rate schedules for non-convex optimization, 2020.
\newblock URL \url{https://api.semanticscholar.org/CorpusID:235266276}.

\bibitem[Defazio and Gower(2020)]{Defazio2020ThePO}
Aaron Defazio and Robert~Mansel Gower.
\newblock The power of factorial powers: New parameter settings for (stochastic) optimization.
\newblock In \emph{Asian Conference on Machine Learning}, 2020.
\newblock URL \url{https://api.semanticscholar.org/CorpusID:224815911}.

\bibitem[Defazio et~al.(2024)Defazio, Yang, Mehta, Mishchenko, Khaled, and Cutkosky]{defazio2024road}
Aaron Defazio, Xingyu Yang, Harsh Mehta, Konstantin Mishchenko, Ahmed Khaled, and Ashok Cutkosky.
\newblock The road less scheduled, 2024.

\bibitem[Defazio et~al.(2025)Defazio, Mishchenko, Raman, Shi, and Xiao]{gpa}
Aaron Defazio, Konstantin Mishchenko, Parameswaran Raman, Hao-Jun~Michael Shi, and Lin Xiao.
\newblock Smoothing diloco with primal averaging for faster training of llms.
\newblock \emph{arXiv preprint arXiv:2512.17131}, 2025.

\bibitem[Deng et~al.(2025)Deng, Wu, Ju, Lin, Liang, and Shroff]{dengiclrtheory}
Junze Deng, Qinhang Wu, Peizhong Ju, Sen Lin, Yingbin Liang, and Ness~B. Shroff.
\newblock Unlocking the power of rehearsal in continual learning: A theoretical perspective.
\newblock \emph{ArXiv}, abs/2506.00205, 2025.
\newblock URL \url{https://api.semanticscholar.org/CorpusID:279074801}.

\bibitem[Freedman(1975)]{ontail}
David~A Freedman.
\newblock On tail probabilities for martingales.
\newblock \emph{the Annals of Probability}, pages 100--118, 1975.

\bibitem[French(1999)]{french1999catastrophic}
Robert~M French.
\newblock Catastrophic forgetting in connectionist networks.
\newblock \emph{Trends in cognitive sciences}, 3\penalty0 (4):\penalty0 128--135, 1999.

\bibitem[Guo et~al.(2022)Guo, Liu, and Zhao]{ocm}
Yiduo Guo, Bing Liu, and Dongyan Zhao.
\newblock Online continual learning through mutual information maximization.
\newblock In \emph{Proceedings of the 39th International Conference on Machine Learning}, volume 162 of \emph{Proceedings of Machine Learning Research}, pages 8109--8126, 17--23 Jul 2022.

\bibitem[Guo et~al.(2023)Guo, Liu, and Zhao]{GSA}
Yiduo Guo, Bing Liu, and Dongyan Zhao.
\newblock Dealing with cross-task class discrimination in online continual learning.
\newblock \emph{2023 IEEE/CVF Conference on Computer Vision and Pattern Recognition (CVPR)}, pages 11878--11887, 2023.
\newblock URL \url{https://api.semanticscholar.org/CorpusID:258865951}.

\bibitem[Gupta et~al.(2020)Gupta, Yadav, and Paull]{lamaml}
Gunshi Gupta, Karmesh Yadav, and Liam Paull.
\newblock La-maml: Look-ahead meta learning for continual learning.
\newblock \emph{arXiv preprint arXiv:2007.13904}, 2020.

\bibitem[Hazan(2016)]{OCO2}
Elad Hazan.
\newblock Introduction to online convex optimization.
\newblock \emph{Foundations and Trends in Optimization}, 2\penalty0 (3-4):\penalty0 157--325, 2016.

\bibitem[He et~al.(2020)He, Mao, Shao, and Zhu]{He2020IncrementalLI}
Jiangpeng He, Runyu Mao, Zeman Shao, and Fengqing Zhu.
\newblock Incremental learning in online scenario.
\newblock In \emph{Proceedings of the IEEE/CVF conference on computer vision and pattern recognition}, pages 13926--13935, 2020.

\bibitem[He et~al.(2016)He, Zhang, Ren, and Sun]{resnet}
Kaiming He, Xiangyu Zhang, Shaoqing Ren, and Jian Sun.
\newblock Deep residual learning for image recognition.
\newblock In \emph{Proceedings of the IEEE conference on computer vision and pattern recognition}, pages 770--778, 2016.

\bibitem[Hoi et~al.(2021)Hoi, Sahoo, Lu, and Zhao]{OnlineSurvey}
Steven~CH Hoi, Doyen Sahoo, Jing Lu, and Peilin Zhao.
\newblock Online learning: A comprehensive survey.
\newblock \emph{Neurocomputing}, 459:\penalty0 249--289, 2021.

\bibitem[Husz{\'a}r(2017)]{Huszr2017NoteOT}
Ferenc Husz{\'a}r.
\newblock Note on the quadratic penalties in elastic weight consolidation.
\newblock \emph{Proceedings of the National Academy of Sciences}, 115:\penalty0 E2496 -- E2497, 2017.
\newblock URL \url{https://api.semanticscholar.org/CorpusID:3415694}.

\bibitem[Izmailov et~al.(2018)Izmailov, Podoprikhin, Garipov, Vetrov, and Wilson]{izmailov2018averaging}
Pavel Izmailov, Dmitrii Podoprikhin, Timur Garipov, Dmitry Vetrov, and Andrew~Gordon Wilson.
\newblock Averaging weights leads to wider optima and better generalization.
\newblock In \emph{Proceedings of the Thirty-Fourth Conference on Uncertainty in Artificial Intelligence}, 2018.
\newblock URL \url{https://arxiv.org/abs/1803.05407}.

\bibitem[Kingma and Ba(2014)]{Adam}
Diederik~P. Kingma and Jimmy Ba.
\newblock Adam: A method for stochastic optimization.
\newblock \emph{CoRR}, abs/1412.6980, 2014.
\newblock URL \url{https://api.semanticscholar.org/CorpusID:6628106}.

\bibitem[Kirkpatrick et~al.(2016)Kirkpatrick, Pascanu, Rabinowitz, Veness, Desjardins, Rusu, Milan, Quan, Ramalho, Grabska-Barwinska, Hassabis, Clopath, Kumaran, and Hadsell]{Kirkpatrick2016OvercomingCF}
James Kirkpatrick, Razvan Pascanu, Neil~C. Rabinowitz, Joel Veness, Guillaume Desjardins, Andrei~A. Rusu, Kieran Milan, John Quan, Tiago Ramalho, Agnieszka Grabska-Barwinska, Demis Hassabis, Claudia Clopath, Dharshan Kumaran, and Raia Hadsell.
\newblock Overcoming catastrophic forgetting in neural networks.
\newblock \emph{Proceedings of the National Academy of Sciences}, 114:\penalty0 3521 -- 3526, 2016.
\newblock URL \url{https://api.semanticscholar.org/CorpusID:4704285}.

\bibitem[Krizhevsky(2009)]{cifar100}
Alex Krizhevsky.
\newblock Learning multiple layers of features from tiny images.
\newblock 2009.
\newblock URL \url{https://api.semanticscholar.org/CorpusID:18268744}.

\bibitem[Lan(2012)]{lan2012optimal}
Guanghui Lan.
\newblock An optimal method for stochastic composite optimization.
\newblock \emph{Mathematical Programming}, 133\penalty0 (1):\penalty0 365--397, 2012.

\bibitem[Lange and Tuytelaars(2020)]{cope}
Matthias~De Lange and Tinne Tuytelaars.
\newblock Continual prototype evolution: Learning online from non-stationary data streams.
\newblock \emph{2021 IEEE/CVF International Conference on Computer Vision (ICCV)}, pages 8230--8239, 2020.
\newblock URL \url{https://api.semanticscholar.org/CorpusID:221447946}.

\bibitem[Le and Yang(2015)]{tiny-imagenet}
Yann Le and Xuan Yang.
\newblock Tiny imagenet visual recognition challenge.
\newblock \emph{CS 231N}, 7\penalty0 (7):\penalty0 3, 2015.

\bibitem[Li et~al.(2024)Li, Lin, Duan, Liang, and Shroff]{liniclrtheory}
Hongbo Li, Sen Lin, Lingjie Duan, Yingbin Liang, and Ness~B Shroff.
\newblock Theory on mixture-of-experts in continual learning.
\newblock \emph{arXiv preprint arXiv:2406.16437}, 2024.

\bibitem[Liang and Li(2024)]{inflora}
Yan-Shuo Liang and Wu-Jun Li.
\newblock Inflora: Interference-free low-rank adaptation for continual learning.
\newblock In \emph{Proceedings of the IEEE/CVF Conference on Computer Vision and Pattern Recognition}, pages 23638--23647, 2024.

\bibitem[Lin et~al.(2022)Lin, Yang, Fan, and Zhang]{lin2022trgp}
Sen Lin, Li~Yang, Deliang Fan, and Junshan Zhang.
\newblock Trgp: Trust region gradient projection for continual learning.
\newblock In \emph{The Tenth International Conference on Learning Representations}, 2022.

\bibitem[Lin et~al.(2023)Lin, Ju, Liang, and Shroff]{linicmltheory}
Sen Lin, Peizhong Ju, Yingbin Liang, and Ness Shroff.
\newblock Theory on forgetting and generalization of continual learning.
\newblock In \emph{International Conference on Machine Learning}, pages 21078--21100. PMLR, 2023.

\bibitem[Lin et~al.(2021)Lin, Shi, Pathak, and Ramanan]{clear100}
Zhiqiu Lin, Jia Shi, Deepak Pathak, and Deva Ramanan.
\newblock The clear benchmark: Continual learning on real-world imagery.
\newblock In \emph{Thirty-fifth Conference on Neural Information Processing Systems Datasets and Benchmarks Track}, 2021.

\bibitem[Liu et~al.(2024)Liu, Yang, Li, Clifton, and Ghanem]{s6mod}
Sihao Liu, Yibo Yang, Xiaojie Li, David~A Clifton, and Bernard Ghanem.
\newblock Enhancing online continual learning with plug-and-play state space model and class-conditional mixture of discretization.
\newblock \emph{arXiv preprint arXiv:2412.18177}, 2024.

\bibitem[Lopez-Paz and Ranzato(2017)]{gem}
David Lopez-Paz and Marc'Aurelio Ranzato.
\newblock Gradient episodic memory for continual learning.
\newblock In \emph{Neural Information Processing Systems}, 2017.
\newblock URL \url{https://api.semanticscholar.org/CorpusID:37308416}.

\bibitem[Loshchilov and Hutter(2017)]{AdamW}
Ilya Loshchilov and Frank Hutter.
\newblock Fixing weight decay regularization in adam.
\newblock \emph{ArXiv}, abs/1711.05101, 2017.
\newblock URL \url{https://api.semanticscholar.org/CorpusID:3312944}.

\bibitem[Mai et~al.(2021)Mai, Li, Kim, and Sanner]{SCR}
Zheda Mai, Ruiwen Li, Hyunwoo~J. Kim, and Scott Sanner.
\newblock Supervised contrastive replay: Revisiting the nearest class mean classifier in online class-incremental continual learning.
\newblock \emph{2021 IEEE/CVF Conference on Computer Vision and Pattern Recognition Workshops (CVPRW)}, pages 3584--3594, 2021.
\newblock URL \url{https://api.semanticscholar.org/CorpusID:232352581}.

\bibitem[Mai et~al.(2022)Mai, Li, Jeong, Quispe, Kim, and Sanner]{MAI202228}
Zheda Mai, Ruiwen Li, Jihwan Jeong, David Quispe, Hyunwoo Kim, and Scott Sanner.
\newblock Online continual learning in image classification: An empirical survey.
\newblock \emph{Neurocomputing}, 469:\penalty0 28--51, 2022.
\newblock ISSN 0925-2312.
\newblock \doi{https://doi.org/10.1016/j.neucom.2021.10.021}.
\newblock URL \url{https://www.sciencedirect.com/science/article/pii/S0925231221014995}.

\bibitem[Marouf et~al.(2024)Marouf, Roy, Tartaglione, and Lathuili{\`e}re]{marouf2024weighted}
Imad~Eddine Marouf, Subhankar Roy, Enzo Tartaglione, and St{\'e}phane Lathuili{\`e}re.
\newblock Weighted ensemble models are strong continual learners.
\newblock In \emph{European Conference on Computer Vision (ECCV)}, 2024.
\newblock URL \url{https://www.ecva.net/papers/eccv_2024/papers_ECCV/html/9002_ECCV_2024_paper.php}.

\bibitem[Mermillod et~al.(2013)Mermillod, Bugaiska, and Bonin]{mermillod2013stability}
Martial Mermillod, Aur{\'e}lia Bugaiska, and Patrick Bonin.
\newblock The stability-plasticity dilemma: Investigating the continuum from catastrophic forgetting to age-limited learning effects, 2013.

\bibitem[Nesterov and Shikhman(2015)]{primal}
Yu~Nesterov and Vladimir Shikhman.
\newblock Quasi-monotone subgradient methods for nonsmooth convex minimization.
\newblock \emph{Journal of Optimization Theory and Applications}, 165\penalty0 (3):\penalty0 917--940, 2015.

\bibitem[Nesterov(1983)]{nesterov1983method}
Yurii Nesterov.
\newblock A method for solving the convex programming problem with convergence rate o (1/k2).
\newblock In \emph{Dokl akad nauk Sssr}, volume 269, page 543, 1983.

\bibitem[Nesterov et~al.(2018)]{nesterov2018lectures}
Yurii Nesterov et~al.
\newblock \emph{Lectures on convex optimization}, volume 137.
\newblock Springer, 2018.

\bibitem[Ostapenko et~al.(2019)Ostapenko, Puscas, Klein, J{\"a}hnichen, and Nabi]{Ostapenko2019LearningTR}
Oleksiy Ostapenko, Mihai~Marian Puscas, Tassilo Klein, Patrick J{\"a}hnichen, and Moin Nabi.
\newblock Learning to remember: A synaptic plasticity driven framework for continual learning.
\newblock \emph{2019 IEEE/CVF Conference on Computer Vision and Pattern Recognition (CVPR)}, pages 11313--11321, 2019.
\newblock URL \url{https://api.semanticscholar.org/CorpusID:102353035}.

\bibitem[Polyak(1990)]{polyak}
Boris~T Polyak.
\newblock New stochastic approximation type procedures.
\newblock \emph{Automat. i Telemekh}, 7\penalty0 (98-107):\penalty0 2, 1990.

\bibitem[Riemer et~al.(2019)Riemer, Cases, Ajemian, Liu, Rish, Tu, and Tesauro]{MER}
Matthew Riemer, Ignacio Cases, Robert Ajemian, Miao Liu, Irina Rish, Yuhai Tu, and Gerald Tesauro.
\newblock Learning to learn without forgetting by maximizing transfer and minimizing interference.
\newblock In \emph{In International Conference on Learning Representations (ICLR)}, 2019.

\bibitem[Rolnick et~al.(2018)Rolnick, Ahuja, Schwarz, Lillicrap, and Wayne]{Rolnick2018ExperienceRF}
David Rolnick, Arun Ahuja, Jonathan Schwarz, Timothy~P. Lillicrap, and Greg Wayne.
\newblock Experience replay for continual learning.
\newblock In \emph{Neural Information Processing Systems}, 2018.
\newblock URL \url{https://api.semanticscholar.org/CorpusID:53860287}.

\bibitem[Ruder(2016)]{SGD}
Sebastian Ruder.
\newblock An overview of gradient descent optimization algorithms.
\newblock \emph{ArXiv}, abs/1609.04747, 2016.
\newblock URL \url{https://api.semanticscholar.org/CorpusID:17485266}.

\bibitem[Ruppert(1988)]{Ruppert}
David Ruppert.
\newblock Efficient estimations from a slowly convergent robbins-monro process.
\newblock 1988.
\newblock URL \url{https://api.semanticscholar.org/CorpusID:108279905}.

\bibitem[Russakovsky et~al.(2014)Russakovsky, Deng, Su, Krause, Satheesh, Ma, Huang, Karpathy, Khosla, Bernstein, Berg, and Fei{-}Fei]{imagenet1k}
Olga Russakovsky, Jia Deng, Hao Su, Jonathan Krause, Sanjeev Satheesh, Sean Ma, Zhiheng Huang, Andrej Karpathy, Aditya Khosla, Michael~S. Bernstein, Alexander~C. Berg, and Li~Fei{-}Fei.
\newblock Imagenet large scale visual recognition challenge.
\newblock \emph{CoRR}, abs/1409.0575, 2014.
\newblock URL \url{http://arxiv.org/abs/1409.0575}.

\bibitem[Saha et~al.(2021)Saha, Garg, and Roy]{GPM}
Gobinda Saha, Isha Garg, and Kaushik Roy.
\newblock Gradient projection memory for continual learning.
\newblock \emph{arXiv preprint arXiv:2103.09762}, 2021.

\bibitem[Shalev-Shwartz(2025)]{oco}
Shai Shalev-Shwartz.
\newblock Online learning and online convex optimization.
\newblock \emph{Foundations and Trends{\textregistered} in Machine Learning}, 4\penalty0 (2):\penalty0 107--194, 2025.

\bibitem[Shamir and Zhang(2012)]{Shamir2012StochasticGD}
Ohad Shamir and Tong Zhang.
\newblock Stochastic gradient descent for non-smooth optimization: Convergence results and optimal averaging schemes.
\newblock \emph{ArXiv}, abs/1212.1824, 2012.
\newblock URL \url{https://api.semanticscholar.org/CorpusID:1217704}.

\bibitem[Soutif-Cormerais et~al.(2023{\natexlab{a}})Soutif-Cormerais, Carta, Cossu, Hurtado, Lomonaco, Van~de Weijer, and Hemati]{soutif2023comprehensive}
Albin Soutif-Cormerais, Antonio Carta, Andrea Cossu, Julio Hurtado, Vincenzo Lomonaco, Joost Van~de Weijer, and Hamed Hemati.
\newblock A comprehensive empirical evaluation on online continual learning.
\newblock In \emph{Proceedings of the IEEE/CVF International Conference on Computer Vision}, pages 3518--3528, 2023{\natexlab{a}}.

\bibitem[Soutif-Cormerais et~al.(2023{\natexlab{b}})Soutif-Cormerais, Carta, and Van~de Weijer]{soutifcormerais2023improving}
Albin Soutif-Cormerais, Antonio Carta, and Joost Van~de Weijer.
\newblock Improving online continual learning performance and stability with temporal ensembles.
\newblock In \emph{Conference on Lifelong Learning Agents}, pages 828--845. PMLR, 2023{\natexlab{b}}.

\bibitem[van~de Ven and Tolias(2019)]{vandeVen2019ThreeSF}
Gido~M. van~de Ven and Andreas~Savas Tolias.
\newblock Three scenarios for continual learning.
\newblock \emph{ArXiv}, abs/1904.07734, 2019.
\newblock URL \url{https://api.semanticscholar.org/CorpusID:119309522}.

\bibitem[Wang et~al.(2021{\natexlab{a}})Wang, Zhang, Jia, Li, Bao, Ma, Zhu, and Zhong]{afec}
Liyuan Wang, Mingtian Zhang, Zhongfan Jia, Qian Li, Chenglong Bao, Kaisheng Ma, Jun Zhu, and Yi~Zhong.
\newblock Afec: Active forgetting of negative transfer in continual learning.
\newblock \emph{arXiv preprint arXiv:2110.12187}, 2021{\natexlab{a}}.

\bibitem[Wang et~al.(2023{\natexlab{a}})Wang, Xie, Zhang, Huang, Su, and Zhu]{hidelora}
Liyuan Wang, Jingyi Xie, Xingxing Zhang, Mingyi Huang, Hang Su, and Jun Zhu.
\newblock Hierarchical decomposition of prompt-based continual learning: Rethinking obscured sub-optimality.
\newblock \emph{Advances in Neural Information Processing Systems}, 2023{\natexlab{a}}.

\bibitem[Wang et~al.(2024{\natexlab{a}})Wang, Zhang, Su, and Zhu]{CLsurveyWang}
Liyuan Wang, Xingxing Zhang, Hang Su, and Jun Zhu.
\newblock A comprehensive survey of continual learning: Theory, method and application.
\newblock \emph{IEEE TPAMI}, 46\penalty0 (8):\penalty0 5362--5383, 2024{\natexlab{a}}.
\newblock \doi{10.1109/TPAMI.2024.3367329}.

\bibitem[Wang et~al.(2024{\natexlab{b}})Wang, Michel, Xiao, and Yamasaki]{ccldc}
Maorong Wang, Nicolas Michel, Ling Xiao, and Toshihiko Yamasaki.
\newblock Improving plasticity in online continual learning via collaborative learning.
\newblock In \emph{Proceedings of the IEEE/CVF Conference on Computer Vision and Pattern Recognition (CVPR)}, pages 23460--23469, June 2024{\natexlab{b}}.

\bibitem[Wang et~al.(2023{\natexlab{b}})Wang, Wang, Wu, Jia, and Meng]{wangcba}
Quanziang Wang, Renzhen Wang, Yichen Wu, Xixi Jia, and Deyu Meng.
\newblock Cba: Improving online continual learning via continual bias adaptor.
\newblock In \emph{ICCV}, 2023{\natexlab{b}}.

\bibitem[Wang et~al.(2024{\natexlab{c}})Wang, Wang, Li, Wei, Wang, Ma, Zheng, and Meng]{WangTMM}
Quanziang Wang, Renzhen Wang, Yuexiang Li, Dong Wei, Hong Wang, Kai Ma, Yefeng Zheng, and Deyu Meng.
\newblock Relational experience replay: Continual learning by adaptively tuning task-wise relationship.
\newblock \emph{IEEE TMM}, 26:\penalty0 9683--9698, 2024{\natexlab{c}}.
\newblock \doi{10.1109/TMM.2024.3397048}.

\bibitem[Wang et~al.(2021{\natexlab{b}})Wang, Li, Sun, and Xu]{Wang_2021_CVPR}
Shipeng Wang, Xiaorong Li, Jian Sun, and Zongben Xu.
\newblock Training networks in null space of feature covariance for continual learning.
\newblock In \emph{CVPR}, pages 184--193, 2021{\natexlab{b}}.

\bibitem[Wei et~al.(2023)Wei, Ye, Huang, Zhang, and Shan]{onpro}
Yujie Wei, Jiaxin Ye, Zhizhong Huang, Junping Zhang, and Hongming Shan.
\newblock Online prototype learning for online continual learning.
\newblock In \emph{Proceedings of the IEEE/CVF International Conference on Computer Vision}, pages 18764--18774, 2023.

\bibitem[Wu et~al.(2024{\natexlab{a}})Wu, Huang, Wang, Meng, and Wei]{wu2023meta}
Yichen Wu, Long-Kai Huang, Renzhen Wang, Deyu Meng, and Ying Wei.
\newblock Meta continual learning revisited: Implicitly enhancing online hessian approximation via variance reduction.
\newblock In \emph{ICLR}, 2024{\natexlab{a}}.

\bibitem[Wu et~al.(2024{\natexlab{b}})Wu, Wang, Zhao, Zheng, Wei, and Huang]{wumitigating}
Yichen Wu, Hong Wang, Peilin Zhao, Yefeng Zheng, Ying Wei, and Long-Kai Huang.
\newblock Mitigating catastrophic forgetting in online continual learning by modeling previous task interrelations via pareto optimization.
\newblock In \emph{ICML}, 2024{\natexlab{b}}.

\bibitem[Wu et~al.(2025)Wu, Piao, Huang, Wang, Li, Pfister, Meng, Ma, and Wei]{wu2025sdlora}
Yichen Wu, Hongming Piao, Long-Kai Huang, Renzhen Wang, Wanhua Li, Hanspeter Pfister, Deyu Meng, Kede Ma, and Ying Wei.
\newblock Sd-lora: Scalable decoupled low-rank adaptation for class incremental learning.
\newblock \emph{arXiv preprint arXiv:2501.13198}, 2025.

\bibitem[Yan et~al.(2024)Yan, Wang, Ma, and Zhong]{yan2024orchestrate}
Hongwei Yan, Liyuan Wang, Kaisheng Ma, and Yi~Zhong.
\newblock Orchestrate latent expertise: Advancing online continual learning with multi-level supervision and reverse self-distillation.
\newblock In \emph{Proceedings of the IEEE/CVF Conference on Computer Vision and Pattern Recognition}, pages 23670--23680, 2024.

\bibitem[Yan et~al.(2021)Yan, Xie, and He]{Yan2021DERDE}
Shipeng Yan, Jiangwei Xie, and Xuming He.
\newblock Der: Dynamically expandable representation for class incremental learning.
\newblock \emph{2021 IEEE/CVF Conference on Computer Vision and Pattern Recognition (CVPR)}, pages 3013--3022, 2021.
\newblock URL \url{https://api.semanticscholar.org/CorpusID:232427923}.

\bibitem[Yoo et~al.(2024)Yoo, Liu, Wood, and Pleiss]{LPR}
Jason Yoo, Yunpeng Liu, Frank Wood, and Geoff Pleiss.
\newblock Layerwise proximal replay: A proximal point method for online continual learning.
\newblock \emph{arXiv preprint arXiv:2402.09542}, 2024.

\bibitem[Zenke et~al.(2017)Zenke, Poole, and Ganguli]{SI}
Friedemann Zenke, Ben Poole, and Surya Ganguli.
\newblock Continual learning through synaptic intelligence.
\newblock In \emph{International conference on machine learning}, pages 3987--3995. Pmlr, 2017.

\bibitem[Zhang et~al.(2019)Zhang, Lucas, Hinton, and Ba]{lookahead}
Michael~R Zhang, James Lucas, Geoffrey Hinton, and Jimmy Ba.
\newblock Lookahead optimizer: k steps forward, 1 step back.
\newblock \emph{arXiv preprint arXiv:1907.08610}, 2019.

\bibitem[Zhou et~al.(2024)Zhou, Wang, Qi, Ye, Zhan, and Liu]{CLsurveyZhou}
Da-Wei Zhou, Qi-Wei Wang, Zhi-Hong Qi, Han-Jia Ye, De-Chuan Zhan, and Ziwei Liu.
\newblock Class-incremental learning: A survey.
\newblock \emph{IEEE TPAMI}, 46\penalty0 (12):\penalty0 9851--9873, 2024.
\newblock \doi{10.1109/TPAMI.2024.3429383}.

\bibitem[Zhou et~al.(2022)Zhou, Wang, Shu, Zhao, and Meng]{zhoudbn}
Minghao Zhou, Quanziang Wang, Jun Shu, Qian Zhao, and Deyu Meng.
\newblock Diagnosing batch normalization in class incremental learning.
\newblock \emph{arXiv preprint arXiv:2202.08025}, 2022.

\end{thebibliography}

\newpage
\appendix

\crefname{section}{Appendix}{Appendices}
\Crefname{section}{Appendix}{Appendices}

\section*{Appendix}

\addtocontents{toc}{\protect\setcounter{tocdepth}{2}}
\begingroup
\renewcommand{\contentsname}{Appendix contents}
\tableofcontents
\endgroup
\clearpage

\section{Proofs}

\subsection{Proof of Proposition 1 (Implicit Hessian)}
\label{ihessian}

We analyze the local dynamics of the SPIN algorithm to verify the effective update rule presented in the main text (Proposition 1). Let $d_t := \phi_t - \theta_t$ denote the displacement between the fast (plastic) and slow (stable) iterates. Using the interpolation $x_t = \theta_t + (1-\beta)d_t$, we rewrite the system dynamics in terms of the variables $(\theta_t, d_t)$:
\begin{equation}
\label{eq:coupled_dynamics}
\begin{aligned}
    \theta_{t+1} - \theta_t &= \alpha(d_t - \gamma g_t), \\
    d_{t+1} &= (1-\alpha)(d_t - \gamma g_t),
\end{aligned}
\end{equation}
where $g_t = \nabla \mathcal{L}_t(x_t)$.

\paragraph{Local Linearization.}
To expose the curvature dependence, we perform a first-order Taylor expansion of the gradient around the slow iterate $\theta_t$. Let $H_t = \nabla^2 \mathcal{L}_t(\theta_t)$ denote the local Hessian. Neglecting higher-order terms, we approximate the gradient at the interpolated point $x_t$ as:
\begin{equation}
    g_t = \nabla \mathcal{L}_t(\theta_t + (1-\beta)d_t) \approx \nabla \mathcal{L}_t(\theta_t) + (1-\beta)H_t d_t.
\end{equation}
Substituting this linear approximation into the evolution equation for $d_t$ in Eq.~\ref{eq:coupled_dynamics} yields:
\begin{equation}
    d_{t+1} \approx (1-\alpha)\left[ d_t - \gamma \left( \nabla \mathcal{L}_t(\theta_t) + (1-\beta)H_t d_t \right) \right].
\end{equation}

\paragraph{Quasi-Steady State Analysis.}
Since SPIN is designed as a two-time-scale system where the fast variate $\phi_t$ (and consequently $d_t$) evolves more rapidly than $\theta_t$, we apply a quasi-steady-state approximation~\cite{qss2}. We solve for the equilibrium displacement $d_t^\ast$ by imposing the stationarity condition $d_{t+1} \approx d_t$:
\begin{equation}
    d_t^\ast = (1-\alpha) d_t^\ast - (1-\alpha)\gamma \nabla \mathcal{L}_t(\theta_t) - (1-\alpha)\gamma(1-\beta)H_t d_t^\ast.
\end{equation}
Rearranging terms to isolate $d_t^\ast$, we obtain:
\begin{equation}
    \left[ \alpha I + (1-\alpha)\gamma(1-\beta)H_t \right] d_t^\ast = -(1-\alpha)\gamma \nabla \mathcal{L}_t(\theta_t).
\end{equation}
Solving for $d_t^\ast$ gives the explicit form:
\begin{equation}
\label{eq:d_steady}
    d_t^\ast = -(1-\alpha)\gamma \left[ \alpha I + (1-\alpha)\gamma(1-\beta)H_t \right]^{-1} \nabla \mathcal{L}_t(\theta_t).
\end{equation}

\paragraph{Effective Update Rule.}
We now substitute the equilibrium displacement $d_t^\ast$ back into the update rule for the slow iterate $\theta_t$ in Eq.~\ref{eq:coupled_dynamics}. Note that the update term is $\alpha(d_t - \gamma g_t)$. From the stationarity condition $d_t \approx (1-\alpha)(d_t - \gamma g_t)$, we have the identity $d_t - \gamma g_t \approx \frac{1}{1-\alpha} d_t$. Therefore, the update becomes:
\begin{equation}
\theta_{t+1} - \theta_t 
\approx \frac{\alpha}{1-\alpha} d_t^\ast
= -\alpha\gamma \left[ \alpha I + (1-\alpha)\gamma(1-\beta)H_t \right]^{-1} 
\nabla \mathcal{L}_t(\theta_t).
\end{equation}
To align with the form presented in the main text, we factorize $\alpha$ from the inverse term. We define the effective regularization coefficient $\rho_t$ as:
\begin{equation}
    \rho_t = \frac{\gamma(1-\alpha)(1-\beta)}{\alpha}.
\end{equation}
Using this definition, we can rewrite the inverse term as:
\begin{equation}
    [\alpha I + (1-\alpha)\gamma(1-\beta)H_t]^{-1}= \frac{1}{\alpha}
      \left[I + \frac{\gamma(1-\alpha)(1-\beta)}{\alpha} H_t \right]^{-1}= \frac{1}{\alpha}[I + \rho_t H_t]^{-1}.
\end{equation}

Substituting this back into the update equation, the $\alpha$ terms cancel out, yielding the final effective update:
\begin{equation}
    \theta_{t+1} \approx \theta_t - \gamma \left[ I + \rho_t H_t \right]^{-1} \nabla \mathcal{L}_t(\theta_t).
\end{equation}

\paragraph{Connection to Preconditioning.}
For sufficiently small step sizes, applying the Neumann series approximation $(I + A)^{-1} \approx I - A$ yields:
\begin{equation}
    \theta_{t+1} \approx \theta_t - \gamma \left( I - \rho_t H_t \right) \nabla \mathcal{L}_t(\theta_t).
\end{equation}
This confirms that SPIN performs a gradient descent step preconditioned by the implicit inverse Hessian approximation $\tilde{H}^{-1} \approx \gamma(I + \rho_t H_t)^{-1}$.

\subsection{Proof of Theorem 1 (Regret Bound)}
\label{rb}

\textbf{Theorem 1.} \textit{Let $\mathcal{X} \subset \mathbb{R}^d$ be a closed convex set with diameter $D$. Under Assumptions (i)--(iv), if we choose $\gamma = \frac{D}{G\sqrt{T}}\sqrt{\frac{\alpha}{2-\alpha}}$ and initialize $\theta_1 = \phi_1$, the regret $R_T$ satisfies the following bound with probability at least $1-\delta$:}
\begin{equation}
    R_T \le \tilde{\mathcal{O}}(\sqrt{T}).
\end{equation}

We consider online convex optimization~\cite{oco,OCO2} with the update rule defined as:
\begin{align}
    x_t &= (1-\beta)\phi_t + \beta\theta_t, \label{eq:interp} \\
    \phi_{t+1} &= \Pi_{\mathcal X}\bigl(\phi_t - \gamma g_t\bigr), \qquad g_t := \nabla \mathcal{L}_t(x_t), \\
    \theta_{t+1} &= \theta_t + \alpha(\phi_{t+1} - \theta_t),
\end{align}
where $\alpha \in (0,1)$ is a constant. We let $\mathcal{F}_t$ denote the filtration generated by the random variables up to round $t$.

For the analysis, we adopt the following standard assumptions:
\begin{enumerate}[label=(\roman*)]
    \item \textbf{Domain:} $\mathcal X$ is a closed convex set with diameter bounded by $D$, i.e., $\sup_{x,y \in \mathcal X} \|x-y\| \le D$.
    \item \textbf{Loss Function:} Each loss function $\mathcal{L}_t$ is convex and $G$-Lipschitz continuous on $\mathcal X$.
    \item \textbf{Stochastic Gradient:} The stochastic gradients $g_t$ are unbiased, $\mathbb E[g_t \mid \mathcal F_t] = \nabla \mathcal{L}_t(x_t)$, and bounded such that $\|g_t\|\le G$ almost surely.
    \item \textbf{Variance:} The conditional variance of the stochastic gradients is bounded by $\sigma^2$, i.e., $\mathbb E[\|g_t - \nabla \mathcal{L}_t(x_t)\|^2 \mid \mathcal F_t] \le \sigma^2$.
\end{enumerate}

We aim to bound the static regret $R_T := \sum_{t=1}^T (\mathcal{L}_t(\theta_t) - \mathcal{L}_t(\theta^\star))$.

\paragraph{1. Regret Decomposition.}
By the convexity of $\mathcal{L}_t$, we have $\mathcal{L}_t(x_t) - \mathcal{L}_t(\theta^\star) \le \langle \nabla \mathcal{L}_t(x_t), x_t - \theta^\star \rangle$~\cite{OCO2}. We decompose the instantaneous regret as follows:
\begin{equation}
\begin{aligned}
\mathcal{L}_t(\theta_t) - \mathcal{L}_t(\theta^\star)
&= \bigl(\mathcal{L}_t(\theta_t) - \mathcal{L}_t(x_t)\bigr)
   + \bigl(\mathcal{L}_t(x_t) - \mathcal{L}_t(\theta^\star)\bigr) \\
&\le G\|\theta_t - x_t\|
   + \langle \nabla \mathcal{L}_t(x_t), x_t - \theta^\star \rangle .
\end{aligned}
\end{equation}

Using the definition of $x_t$ in Eq.~\ref{eq:interp}, we have $\|\theta_t - x_t\| = (1-\beta)\|\theta_t - \phi_t\|$. Next, we introduce the martingale difference sequence $\xi_t := \langle \nabla \mathcal{L}_t(x_t) - g_t, x_t - \theta^\star \rangle$. We rewrite the inner product term using the stochastic gradient $g_t$:
\begin{equation}
    \langle \nabla \mathcal{L}_t(x_t), x_t - \theta^\star \rangle = \langle g_t, x_t - \theta^\star \rangle + \xi_t \nonumber = \langle g_t, \phi_t - \theta^\star \rangle + \langle g_t, x_t - \phi_t \rangle + \xi_t.
\end{equation}
Noting that $x_t - \phi_t = \beta(\theta_t - \phi_t)$ and $\|g_t\| \le G$ (Assumption iii), we bound $\langle g_t, x_t - \phi_t \rangle \le G\beta \|\theta_t - \phi_t\|$. Combining these terms yields:
\begin{equation}
\label{eq:decomp_final}
    \mathcal{L}_t(\theta_t) - \mathcal{L}_t(\theta^\star)
    \le \langle g_t, \phi_t - \theta^\star \rangle + G\|\theta_t - \phi_t\| + \xi_t.
\end{equation}

\paragraph{2. Bounding the Gradient Descent Term.}
The update for $\phi_t$ corresponds to Online Gradient Descent (OGD)~\cite{OCO2}. By the standard analysis of OGD and the non-expansiveness of the projection operator $\Pi_{\mathcal{X}}$, we have:
\begin{equation}
    \langle g_t, \phi_t - \theta^\star \rangle
    \le \frac{\|\phi_t - \theta^\star\|^2 - \|\phi_{t+1} - \theta^\star\|^2}{2\gamma} + \frac{\gamma}{2}\|g_t\|^2.
\end{equation}

\paragraph{3. Bounding the Drift Term.}
Let $\Delta_t = \theta_t - \phi_t$. From the update rule $\theta_{t+1} = (1-\alpha)\theta_t + \alpha\phi_{t+1}$, the recurrence for the difference is:
\begin{equation}
    \Delta_{t+1} = (1-\alpha)\theta_t + \alpha\phi_{t+1} - \phi_{t+1} = (1-\alpha)(\theta_t - \phi_{t+1}).
\end{equation}
Taking norms and applying the triangle inequality with $\|\phi_t - \phi_{t+1}\| \le \gamma G$:
\begin{equation}
    \|\Delta_{t+1}\| \le (1-\alpha)\|\Delta_t\| + (1-\alpha)\gamma G.
\end{equation}
Summing this geometric series over $t=1,\dots,T$ yields:
\begin{equation}
    \sum_{t=1}^T \|\theta_t - \phi_t\| \le \frac{\|\theta_1 - \phi_1\|}{\alpha} + T\frac{1-\alpha}{\alpha}\gamma G.
\end{equation}

\paragraph{4. Concentration Analysis.}
Since $\|x_t - \theta^\star\| \le D$ (Assumption i) and gradients are bounded by $G$, the martingale difference is bounded as $|\xi_t| \le 2GD$. The conditional variance is bounded by $\mathbb{E}[\xi_t^2 \mid \mathcal{F}_t] \le D^2\sigma^2$ (Assumption iv). Applying Freedman's inequality~\cite{ontail}, with probability at least $1-\delta$:
\begin{equation}
    \sum_{t=1}^T \xi_t \le D\sigma\sqrt{2T\ln\tfrac{1}{\delta}} + \frac{2GD}{3}\ln\tfrac{1}{\delta}.
\end{equation}

\paragraph{5. Final Bound.}
Summing Eq.~\ref{eq:decomp_final} over $t=1$ to $T$ and substituting the bounds derived above:
\begin{equation}
R_T
\le
\frac{G\|\theta_1-\phi_1\|}{\alpha}
+
\left[
\frac{D^2}{2\gamma}
+
\frac{\gamma T G^2}{2}
+
\gamma T G^2 \frac{1-\alpha}{\alpha}
\right]
+
\sum_{t=1}^{T}\xi_t .
\end{equation}

Substituting the optimal step size $\gamma^\star = \frac{D}{G\sqrt{T}}\sqrt{\frac{\alpha}{2-\alpha}}$, the term in brackets simplifies to $DG\sqrt{T}\sqrt{\frac{2-\alpha}{\alpha}}$. Thus, with probability $\ge 1-\delta$:
\begin{equation}
    R_T \le\ 
    \frac{G\|\theta_1 - \phi_1\|}{\alpha}
    + DG\sqrt{T}\sqrt{\frac{2-\alpha}{\alpha}} \quad
    + D\sigma\sqrt{2T\ln\tfrac{1}{\delta}}
    + \frac{2GD}{3}\ln\tfrac{1}{\delta}.
\end{equation}
Assuming the initialization $\theta_1 = \phi_1$, the first term vanishes, and we obtain the final asymptotic bound:
\begin{equation}
    R_T \le \tilde{\mathcal{O}}(\sqrt{T}).
\end{equation}

\paragraph{Conclusion.}
Theorem 1 establishes that the proposed algorithm achieves sublinear regret matching the optimal $\tilde{\mathcal{O}}(\sqrt{T})$ lower bound for stochastic OCO. The constant $\sqrt{(2-\alpha)/\alpha}$ highlights the trade-off between the memory retention of the slow weights (parameterized by $\alpha$) and the convergence speed.

\section{An Intuitive Understanding of Two Averaging}
\begin{figure}[t]
  \centering
  \begin{subfigure}[t]{0.4\linewidth}
    \centering
    \includegraphics[width=\linewidth]{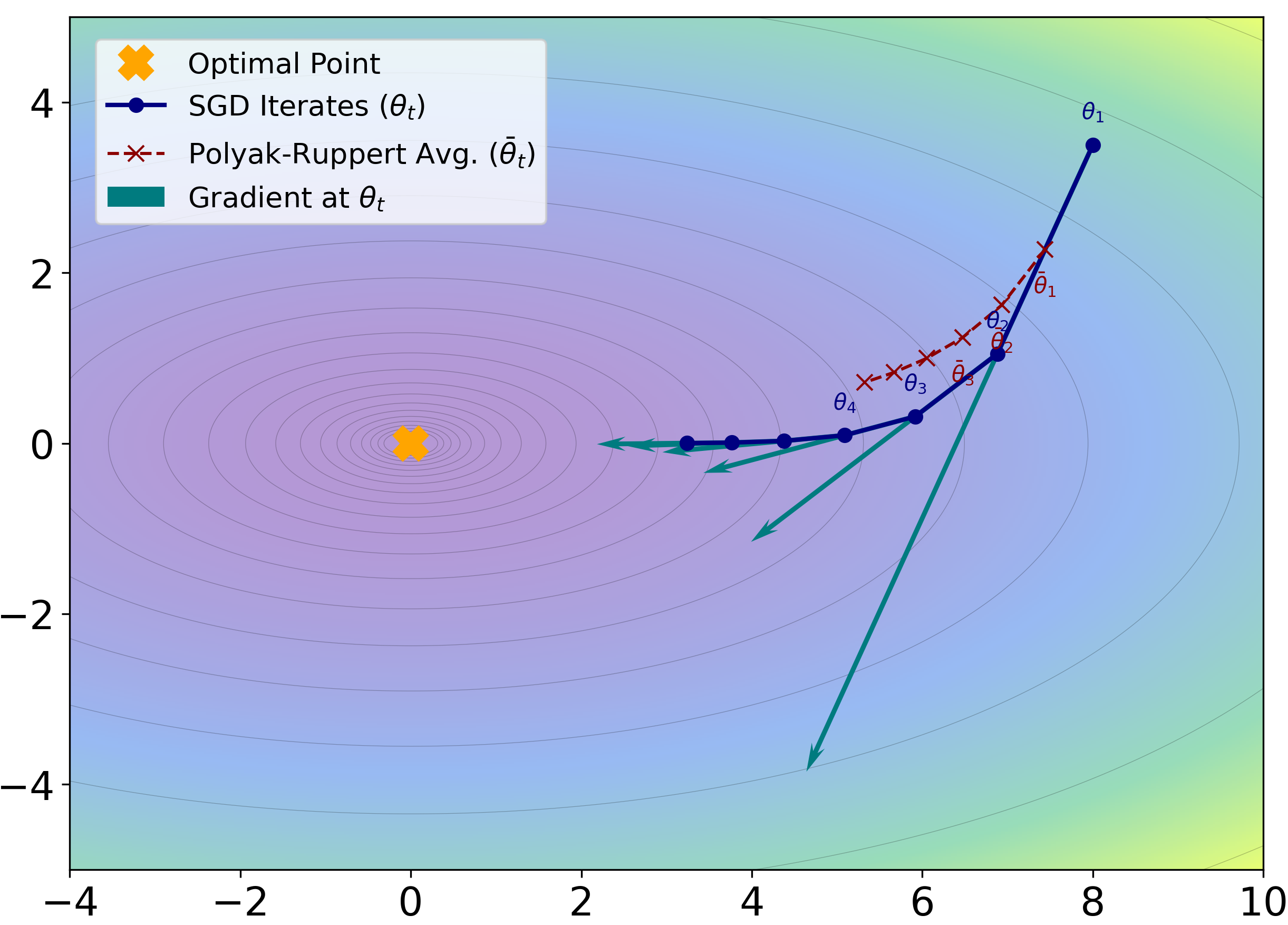}
    \caption{Polyak-Ruppert averaging.}
    \label{fig:sub1}
  \end{subfigure}
  \begin{subfigure}[t]{0.4\linewidth}
    \centering
    \includegraphics[width=\linewidth]{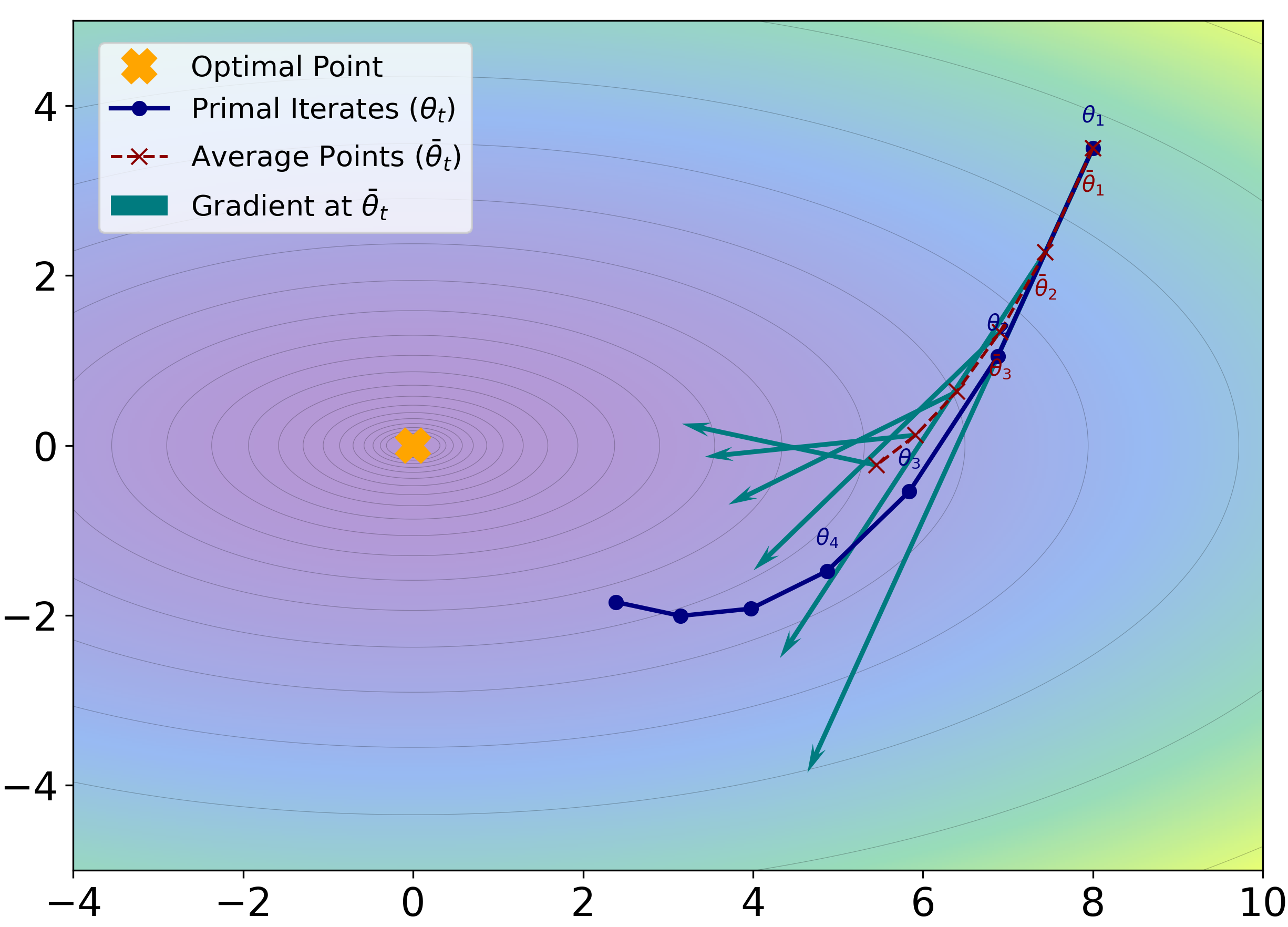}
    \caption{Primal averaging.}
    \label{fig:sub2}
  \end{subfigure}
  \caption{
  Trajectories of Polyak-Ruppert Averaging and Primal Averaging. 
  \textbf{Polyak-Ruppert Averaging} computes the gradient 
  at the current, fast-moving model iterate $\theta_t$: $g_t = \nabla \mathcal{L}(\theta_t)$. 
  The final model is the average of this noisy sequence: 
  $\bar{\theta}_T = \frac{1}{T}\sum_{t=1}^T \theta_t$.
  \textbf{Primal Averaging} computes the gradient 
  at the averaged iterate over the current trajectory: $g_t = \nabla \mathcal{L}(\bar{\theta}_t)$, where 
  $\bar{\theta}_t = \frac{1}{t}\sum_{i=1}^t \theta_i$. 
  This stable gradient is then used to guide the update. 
}
  \label{fig:twoavg}

\end{figure}

We present these two averaging schemes, as shown in \Cref{fig:twoavg}. 

\section{pipeline}
\begin{figure*}[t]
  \centering
  \includegraphics[width=\linewidth, trim=0.2cm 6cm 1.8cm 2cm, clip]{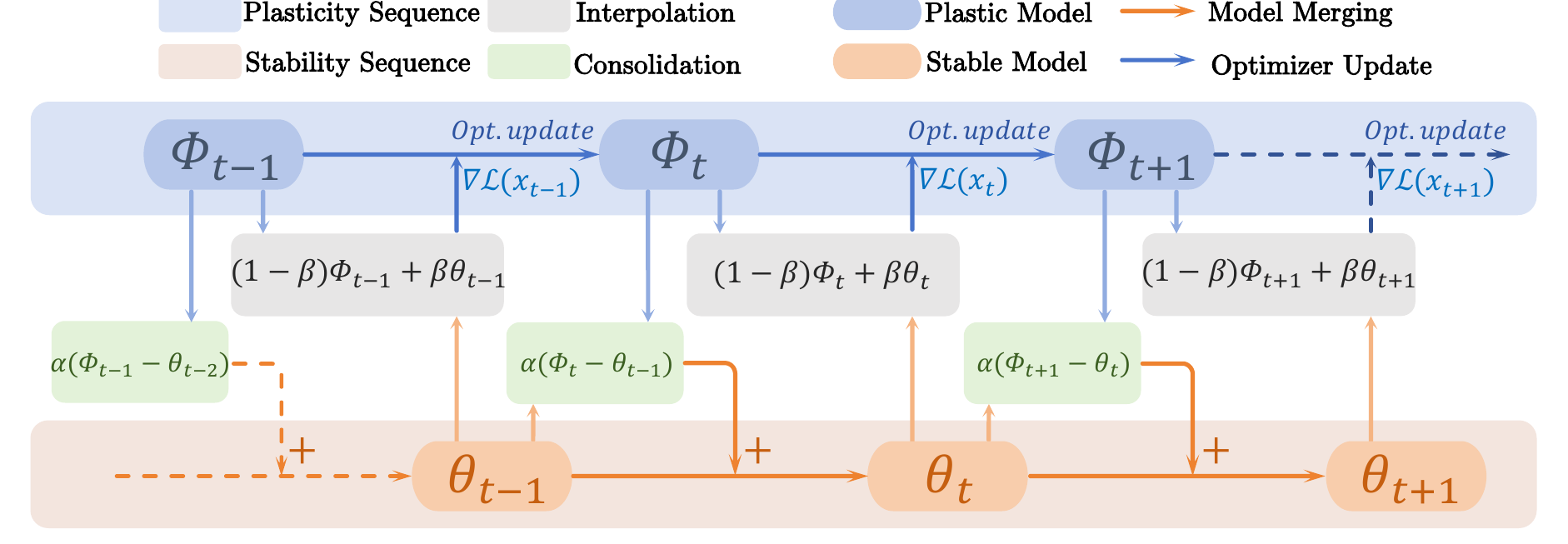}
  \caption{The pipeline of SPIN. The gradient $\nabla \mathcal{L}$ at the model interpolation is obtained from the upstream design of OCL algorithms.}
  \label{fig:pipeline}
\end{figure*}

We present the pipeline of SPIN in~\Cref{fig:pipeline}.

\section{Implementation Details}

\subsection{Dataset Details}
\label{datasetd}
In this work, we experiment with five benchmark datasets in online continual learning setting: CIFAR10~\cite{cifar100}, CIFAR100~\cite{cifar100}, Tiny-Imagenet~\cite{tiny-imagenet}, CLEAR100~\cite{clear100}, and Imagenet1K~\cite{imagenet1k}, which are detailed as follows:

\textbf{Seq-CIFAR10~\cite{cifar100}.}
    Seq-CIFAR10 is a widely used benchmark in Online Continual Learning (OCL), constructed by splitting the CIFAR-10 dataset into sequential tasks. It contains 10 classes and 60,000 RGB images of size $32 \times 32$, and is commonly used to evaluate catastrophic forgetting under class-incremental learning settings.

\textbf{Seq-CIFAR100~\cite{cifar100}.}
    Seq-CIFAR100 is a continual learning benchmark derived from CIFAR-100, consisting of 100 fine-grained categories and 60,000 images. Due to its larger number of classes and more complex task splits, it provides a more challenging setting for evaluating the stability and generalization ability of OCL methods.

\textbf{Tiny-ImageNet~\cite{tiny-imagenet}.}
    Tiny-ImageNet is a reduced version of ImageNet containing 200 classes with $64 \times 64$ resolution images. Its higher visual diversity and semantic complexity make it a popular benchmark for testing representation learning and forgetting resistance in online continual learning scenarios.

\textbf{CLEAR100~\cite{clear100}.}
    CLEAR100 is specifically designed for online continual learning and emphasizes realistic streaming data settings. It includes distribution shifts, sequential data streams, and limited replay constraints, making it suitable for evaluating adaptability and robustness in dynamic learning environments.

\textbf{ImageNet-1K~\cite{imagenet1k}.}
    ImageNet-1K is one of the most influential large-scale image classification datasets, containing 1,000 categories and over one million high-resolution images. In OCL research, it is commonly used to build large-scale class-incremental or streaming benchmarks for evaluating long-term memory retention and continual adaptation.

\subsection{Evaluation Metrics}

We adopt several metrics commonly used in online
continual learning.
Let $a_{i,j}$ denote the accuracy of the model after learning task $i$ on the test data of task $j$.

\paragraph{Average Accuracy (Acc).}
Acc measures the final performance after completing all tasks:
\begin{equation}
\mathrm{Acc} = \frac{1}{T}\sum_{j=1}^T a_{T,j}.
\label{eq:acc}
\end{equation}

\paragraph{Averaged Anytime Accuracy (AAA).}
AAA evaluates the performance throughout the learning process:
\begin{equation}
\mathrm{AAA} = 
\frac{1}{T}\sum_{i=1}^{T}
\left( \frac{1}{i}\sum_{j=1}^{i} a_{i,j} \right).
\label{eq:aaa}
\end{equation}

\paragraph{Learning Accuracy (LA).}
LA quantifies plasticity as the accuracy on each task immediately after it is learned:
\begin{equation}
l_j = a_{j,j},
\end{equation}
and the overall learning accuracy is
\begin{equation}
\mathrm{LA} = \frac{1}{T}\sum_{j=1}^T l_j.
\label{eq:la}
\end{equation}

\paragraph{Average Forgetting (AF).}
AF measures the degradation on old tasks:
\begin{equation}
\mathrm{AF} = \frac{1}{T-1}\sum_{j=1}^{T-1}
\left( \max_{k\le T} a_{k,j} - a_{T,j} \right).
\label{eq:af}
\end{equation}

\paragraph{Relative Forgetting (RF).}
Relative Forgetting measures the proportion of performance degradation on previous tasks.
For the $j$-th task after learning task $k$, it is defined as
\begin{equation}
f_j^k =
\max_{i\in\{1,\dots,k\}}
\left(
1 - \frac{a_j^k}{a_j^i}
\right),
\quad j \le k .
\end{equation}

The overall relative forgetting averaged across all tasks is
\begin{equation}
\mathrm{RF} =
\frac{1}{T}
\sum_{j=1}^{T} f_j^T .
\label{eq:rf}
\end{equation}

\subsection{Training Details}
\label{td}

For all baselines except MOSE~\cite{yan2024orchestrate} and SCR~\cite{SCR}, we follow the
standard experimental configuration used in CCL-DC~\cite{ccldc}. All
methods adopt a randomly initialized ResNet-18 backbone
without pretraining. The model operates in a strict one-pass
online continual learning setting, where each training sample
is observed exactly once. The streaming batch size is fixed
to $10$, and at every iteration a mini-batch of $64$ samples
is drawn from the replay buffer. The memory buffer is
maintained using reservoir sampling, and identical data
augmentation is applied across all baselines to ensure
fairness. 

MOSE and SCR follow the experimental setup of MOSE. Both methods use a nearest-class-mean (NCM) classifier for final prediction. The same online protocol is applied, including a streaming batch size of $10$, a replay batch size of $64$, and reservoir-based memory management. All experiments were conducted using one NVIDIA RTX 4090 GPU.

\begin{table}[htbp]
\centering
\caption{Learning rates for each OCL method and optimizer.}
\label{tab:learning_rates}
\begin{tabular}{lcccc}
\toprule
\textbf{Method} & \textbf{SGD} & \textbf{SPIN-SGD} & \textbf{AdamW} & \textbf{SPIN-AdamW} \\
\midrule
ER & 0.01 & 0.01 & 0.0005 & 0.0005 \\
DER++ & 0.03 & 0.03 & 0.0005 & 0.0005 \\
COPE & 0.03 & 0.03 & 0.0005 & 0.0005 \\
SCR & 0.05 & 0.05 & 0.005 & 0.005 \\
ER-ACE & 0.01 & 0.01 & 0.0005 & 0.0005 \\
OCM & 0.05 & 0.05 & 0.001 & 0.001 \\
OnPro & 0.03 & 0.03 & 0.0005 & 0.0005 \\
GSA & 0.03 & 0.03 & 0.0005 & 0.0005 \\
MOSE & 0.05 & 0.05 & 0.001 & 0.001 \\
CCL-DC & 0.03 & 0.03 & 0.0005 & 0.0005 \\
S6MOD & 0.03 & 0.03 & 0.0005 & 0.0005 \\
\bottomrule
\end{tabular}
\end{table}

\newtext{Learning rates for the 11 OCL methods are summarized in \Cref{tab:learning_rates}.} Note that in most of our experiments, SPIN can replace the existing optimizer without any additional adjustments.

\paragraph{Data Augmentation.}
Data augmentation has been shown to effectively improve the performance of online continual learning (OCL) methods~\cite{derpp,ccldc}. Since different methods may respond differently to augmentation strength, we consider two commonly used augmentation strategies to ensure a fair and comprehensive comparison.

\begin{enumerate}
    \item \textbf{Partial Augmentation Strategy.}
    This strategy adopts relatively weak data augmentation, including random cropping with probability $p=0.5$ and random horizontal flipping with probability $p=0.5$.

    \item \textbf{Full Augmentation Strategy.}
    The full augmentation strategy extends the partial augmentation setting by additionally incorporating color jitter and random grayscale transformations. Specifically, the color jitter parameters are set to $(0.4, 0.4, 0.4, 0.1)$ with probability $p=0.8$, while random grayscale is applied with probability $p=0.2$.
\end{enumerate}

For fair comparison, the models trained with SPIN also adopt the same augmentation strategy in the baseline loss branch. Note that the overall augmentation setting is kept consistent with CCL-DC~\cite{ccldc}. We present hyperparameter searches for some of the methods in ~\Cref{tab:hp_search}.

\begin{table}[t]
\centering
\caption{Hyperparameter for part of baseline methods.}
\label{tab:hp_search}
\begin{tabular}{c c c}
\toprule
\textbf{Method} & \textbf{HP} & \textbf{Values} \\
\midrule

\multirow{5}{*}{ER}
& optimizer & [SGD, AdamW, SPIN, RMSProp, Adam] \\
& lr & [0.1, 0.3, 0.05, 0.01, 0.005, 0.001, 0.0005, 0.0001] \\
& weight decay & [1e-4] \\
& momentum (for SGD) & [0.9] \\
& aug. strat. & [partial, full] \\
\midrule

\multirow{7}{*}{DER++}
& optimizer & [SGD, AdamW, SPIN, RMSProp, Adam] \\
& lr & [0.1, 0.3, 0.05, 0.01, 0.005, 0.001, 0.0005, 0.0001] \\
& weight decay & [0, 1e-4] \\
& momentum (for SGD) & [0, 0.9] \\
& aug. strat. & [partial, full] \\
& alpha & [0.2] \\
& beta & [1.0] \\
\midrule

\multirow{5}{*}{GSA}
& optimizer & [SGD, AdamW, SPIN, RMSProp, Adam] \\
& lr & [0.1, 0.3, 0.05, 0.01, 0.005, 0.001, 0.0005, 0.0001] \\
& weight decay & [1e-4] \\
& momentum (for SGD) & [0.9] \\
& aug. strat. & [partial, full] \\
\bottomrule
\end{tabular}
\end{table}

\section{Performance on Different Architecture}

\begin{table}[t]
\centering
\caption{Optimization performance on the ViT architecture.}
\label{tab:arch}
\begin{tabular}{c|cc}
\toprule
\textbf{Metric} & \textbf{AdamW} & \textbf{SPIN} \\
\midrule
Acc (\%) & 25.89 & \textbf{26.72} \\
AAA (\%) & 33.65 & \textbf{34.98} \\
\bottomrule
\end{tabular}
\end{table}

To further evaluate the robustness of our method across different model architectures, we additionally conduct experiments on a Transformer-based backbone. Specifically, we compare the optimization performance of AdamW and our SPIN under the Vision Transformer (ViT) architecture using the GSA framework. In these experiments, the memory buffer size is set to 2000 and the learning rate is fixed at $0.0001$. The results demonstrate that our method consistently improves performance across different optimization landscapes and network structures, highlighting its generalization capability in online continual learning settings. Note that ResNet-18 is the standard backbone in OCL, while ViT remains largely unexplored.

\section{Analysis of Hyperparameter Robustness And Comparisons with SGD+EMA Baselines.}
\label{hpema}
\begin{figure}[t]
  \centering
  \includegraphics[width=0.6\linewidth]{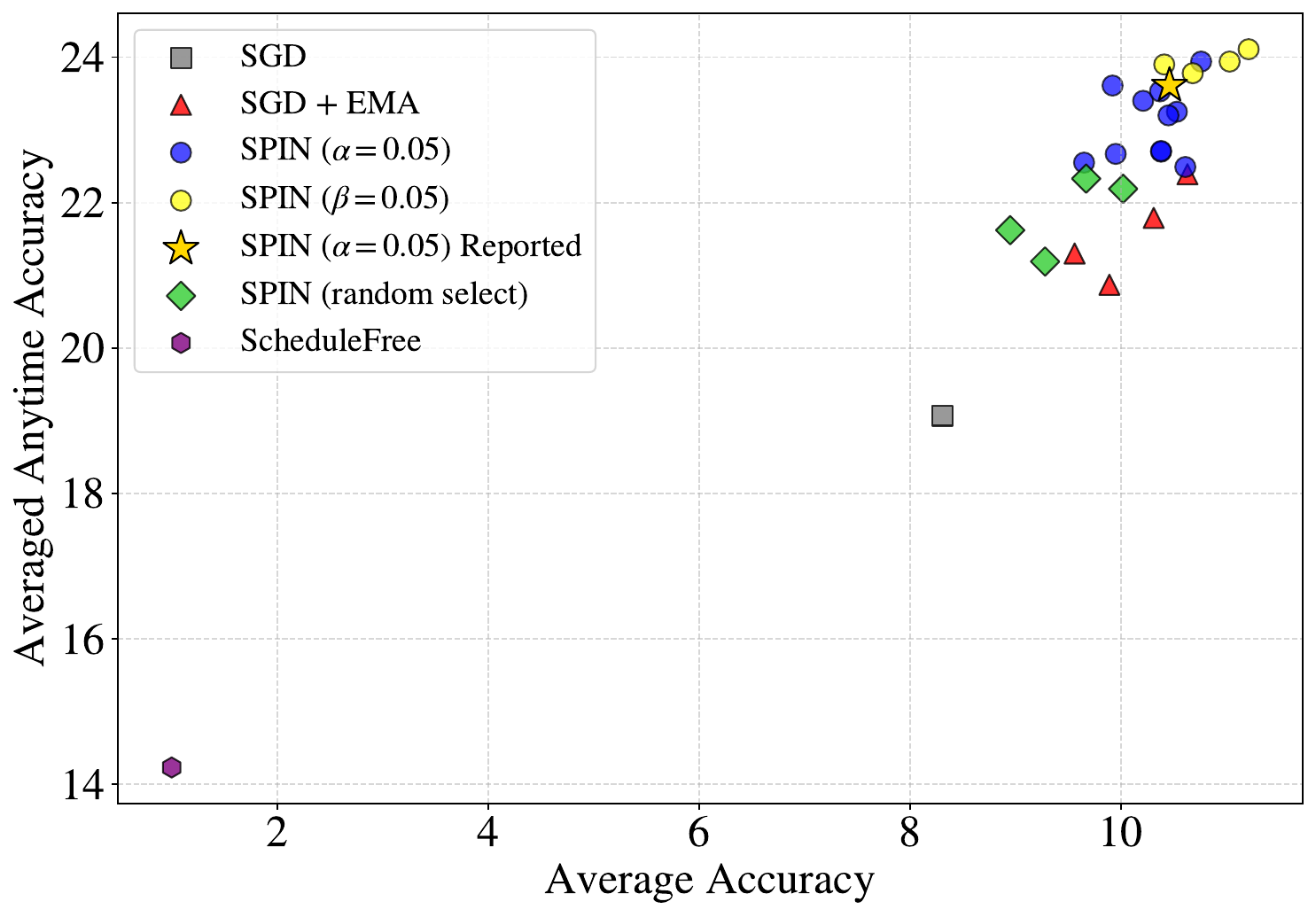}
  \caption{Analysis of hyperparameter robustness and comparisons with different baselines.}
  \label{fig:rr}
\end{figure}

 We evaluate SPIN's sensitivity to $\alpha$ and $\beta$ in \Cref{fig:rr}. Even randomly selected $(\alpha, \beta)$ pairs consistently outperform standard SGD. Specifically, fixing one parameter to an empirical default (\eg, $\alpha=0.05$ or $\beta=0.05$) while tuning the other within $[0.01, 0.99]$ yields peak performance, allowing SPIN to consistently surpass the tuned EMA baseline. This superiority arises because EMA effectively operates as a subset of SPIN. Finally, ScheduleFree performs poorly in this context, as its equal-weighted scheduling implicitly assumes a stationary distribution, which mismatches the strictly non-stationary nature of OCL.

\section{Relation to Multi-Sequence Optimizers}
\label{sec:related_optimizers}

\newtext{We discuss how multi-sequence optimizers choose gradient evaluation points and average iterates. Throughout this section, $z_t$ denotes the base iterate, $x_t$ the averaged iterate, and $y_t$ the gradient evaluation point. The formulas below use gradient-descent updates without weight decay to isolate the averaging mechanisms.}

\paragraph{Lookahead.}
\newtext{Lookahead~\cite{lookahead} performs $k$ base-optimizer steps on fast weights, interpolates the slow weights toward the resulting fast weights, and resets the fast weights to the updated slow weights. With SGD as the base optimizer and $z_{t,0}=x_t$, these updates are}
{
\begin{align}
    z_{t,i+1} &= z_{t,i}-\gamma_{t,i}g_{t,i}(z_{t,i}), \quad i=0,\ldots,k-1, \label{eq:app_lookahead_fast} \\
    x_{t+1} &= (1-\lambda)x_t+\lambda z_{t,k}, \label{eq:app_lookahead_slow} \\
    z_{t+1,0} &= x_{t+1}. \label{eq:app_lookahead_reset}
\end{align}
}
\newtext{Here, $t$ indexes synchronization rounds, $i$ indexes inner steps, $g_{t,i}$ is the stochastic gradient at that step, and $\lambda\in(0,1]$ is the slow-weight interpolation coefficient. Its averaging is periodic; the methods below instead couple their sequences at every step.}

\paragraph{AC-SA form.}
\newtext{For an $L$-smooth convex objective $f$, the accelerated formulation discussed in the main text can be written as~\cite{nesterov1983method,lan2012optimal,defazio2024road}}
{
\begin{align}
    y_t &= (1-c_{t+1})x_t + c_{t+1}z_t, \label{eq:app_acsa_y} \\
    z_{t+1} &= z_t - \frac{t+1}{2L}\nabla f(y_t), \label{eq:app_acsa_z} \\
    x_{t+1} &= (1-c_{t+1})x_t + c_{t+1}z_{t+1}, \label{eq:app_acsa_x}
\end{align}
}
\newtext{where $c_{t+1}=2/(t+2)$. The same time-dependent coefficient controls both interpolations, together with a growing base step size. This coupling is motivated by accelerated convergence for a fixed objective.}

\paragraph{Schedule-Free optimization.}
\newtext{The basic uniform-averaging version of Schedule-Free SGD takes the form~\cite{defazio2024road}}
{
\begin{align}
    y_t &= (1-\beta)z_t + \beta x_t, \label{eq:app_sf_y} \\
    z_{t+1} &= z_t - \gamma_t g_t(y_t), \label{eq:app_sf_z} \\
    x_{t+1} &= \frac{t}{t+1}x_t + \frac{1}{t+1}z_{t+1}, \label{eq:app_sf_x}
\end{align}
}
\newtext{where $g_t(y_t)$ is a stochastic gradient. Unlike the AC-SA form above, the gradient interpolation coefficient is fixed while the averaging coefficient decreases. The framework also supports weighted averages, and nonconvex guarantees have been established for Schedule-Free SGD~\cite{ahn2025general}. In OCL, a decreasing averaging coefficient may reduce responsiveness to newly arriving data; its effect on retention and adaptation requires empirical assessment.}

\paragraph{Generalized Primal Averaging (GPA).}
\newtext{GPA~\cite{gpa} uses two independent interpolation coefficients:}
{
\begin{align}
    y_t &= \mu_y x_t + (1-\mu_y)z_t, \label{eq:app_gpa_y} \\
    z_{t+1} &= z_t - \gamma_t g_t(y_t), \label{eq:app_gpa_z} \\
    x_{t+1} &= \mu_x x_t + (1-\mu_x)z_{t+1}. \label{eq:app_gpa_x}
\end{align}
}
\newtext{Here, $\mu_y$ controls the gradient evaluation point, whereas $\mu_x$ controls the persistence of the averaged sequence. Expanding the last recursion gives}
{
\begin{equation}
    x_{t+1}=\mu_x^{t+1}x_0+(1-\mu_x)\sum_{j=0}^{t}\mu_x^{t-j}z_{j+1}.
    \label{eq:app_gpa_weights}
\end{equation}
}
\newtext{Thus, GPA assigns exponentially decaying weights to past base iterates, with a constant weight on each new iterate. GPA was developed to smooth the periodic updates of single-worker DiLoCo and improve training efficiency. Its fixed averaging coefficient does not automatically inherit the schedule-free property of cumulative averaging.}

\paragraph{Perspective for OCL.}
\newtext{Our focus is the behavior of these dynamics under non-stationary, single-pass learning: how gradient interpolation and averaging balance acquisition of new knowledge with retention of previous tasks. Motivated by the stability–plasticity dilemma, SPIN interprets the fast sequence as adaptation and the slow sequence as progressive consolidation. The curvature interpretation in \Cref{sec4.2} and the accuracy--forgetting analysis in \Cref{ablation} examine these questions in the OCL setting.}

\section{Detailed Experimental Setup for Sec.~4.3}
\label{supp:sec43_setup}

\subsection{Setup for verifying the quasi-steady-state assumption}
To validate the quasi-steady-state (QSS) assumption used in our derivation, we measure
\begin{equation}
    r_t = \frac{\|d_{t+1}\|}{\|d_t\|},
\end{equation}
where $d_t=\phi_t-\theta_t$ is the gap between the plasticity sequence and the stability sequence.

We use the ER method as the base OCL framework and keep the setup fully consistent with the main experiments. The experiment is conducted on Split CIFAR-100, with 10 tasks and 10 classes per task. The backbone is ResNet-18. The stream batch size is 10, the replay batch size is 64, and the replay buffer size is 2000. We use SPIN-AdamW with $\alpha=0.05$ and $\beta=0.05$. All remaining hyperparameters are the same as those used in the corresponding main experiments.

During training, we record $r_t$ at every optimization step. The resulting curve is reported in ~\Cref{fig:emp}a of the main paper.

\subsection{Setup for comparing update norms with Newton's method}

To further validate the curvature-related interpretation of SPIN, we compare the $\ell_2$ norm of its update with that of a Newton-type method.

Because computing Newton updates on CIFAR-100 with a ResNet-18 backbone is computationally too expensive due to the high dimensionality of the model parameters and the cost of Hessian computation, we instead perform this experiment on MNIST. This choice allows efficient computation of second-order information while preserving the purpose of the experiment, namely, comparing update norm behavior.

We train a small multilayer perceptron (MLP) classifier on MNIST using the standard cross-entropy loss. The network consists of two fully connected hidden layers with 256 units each and ReLU activations. The batch size is set to 128, and the model is trained for 20 epochs. 

We compare two optimization methods:
\begin{itemize}
    \item \textbf{SPIN}: our proposed optimizer built on top of SGD.
    \item \textbf{Newton}: a Newton-type method that computes updates using the gradient and the Hessian of the current mini-batch loss.
\end{itemize}

For SPIN, we use a base learning rate of \textbf{0.01}, with hyperparameters $\alpha = 0.05$ and $\beta = 0.05$, which are consistent with the settings used in the main experiments.

At each training step $t$, we compute the parameter update vector $\Delta_t$ produced by each method and record its $\ell_2$ norm $\|\Delta_t\|_2$. For SPIN, $\Delta_t$ denotes the actual update applied by the optimizer to the model parameters. For the Newton method, $\Delta_t$ corresponds to the second-order update computed from the gradient and Hessian.

Both methods use the same data order and training objective to ensure a fair comparison. Since the purpose of this experiment is theoretical validation rather than performance benchmarking, we focus on the evolution of the update norm during training instead of the final classification accuracy. The resulting comparison is shown in ~\cref{fig:emp}b of the main paper.

\clearpage
\section{Algorithm Pseudocode}
\label{pcode}

\begin{algorithm}

\caption{SPIN-SGD for OCL}
\label{alg:spin_sgd}
\DontPrintSemicolon
\KwIn{Task stream $\{\mathcal{D}_t\}_{t=1}^T$; replay buffer capacity $\mathcal{M}$;
      interpolation coefficient $\beta$; stability merging coefficient $\alpha$;
      learning rate $\gamma$; weight decay $\lambda$.}
\KwOut{Final stability parameters $\theta_S$.}

Initialize network parameters $\theta_1$; set $\phi_1 \leftarrow \theta_1$.\;
Initialize replay buffer $\mathcal{M} \leftarrow \emptyset$.\;
Initialize global step counter $s \leftarrow 1$.\;

\For{$t = 1$ \KwTo $T$}{
  \For{each incoming mini-batch $\mathcal{B}^{\text{stream}} \subset \mathcal{D}_t$}{
    Sample replay batch $\mathcal{B}^{\mathcal{M}} \sim \mathcal{M}$ if $\mathcal{M}$ is not empty.\;
    Construct training batch $\mathcal{B}_s \leftarrow \mathcal{B}^{\text{stream}} \cup \mathcal{B}^{\mathcal{M}}$.\;

    Compute interpolated parameters $x_s \leftarrow (1-\beta)\phi_s + \beta\theta_s$.\;

    Compute loss $\mathcal{L}_s$ on $\mathcal{B}_s$ using the upstream OCL method with parameters $x_s$.\;
    Compute stochastic gradient $g_s \leftarrow \nabla \mathcal{L}_s(x_s)$.\;

    \tcp{Plasticity update}
    $\phi_{s+1} \leftarrow \phi_s - \gamma (g_s + \lambda x_s)$.\;

    \tcp{Stability update}
    $\theta_{s+1} \leftarrow \theta_s + \alpha(\phi_{s+1} - \theta_s)$.\;

    Update replay buffer $\mathcal{M}$ with $\mathcal{B}^{\text{stream}}$.\;
    $s \leftarrow s + 1$.\;
  }
}
\Return{$\theta_S$}\;
\end{algorithm}

\end{document}